\documentclass[letterpaper, 10 pt, journal, twoside]{IEEEtran}
\usepackage{graphics} % for pdf, bitmapped graphics files
\usepackage{epsfig} % for postscript graphics files
\usepackage{times} % assumes new font selection scheme installed
 
\usepackage{float}

\let\labelindent\relax
\usepackage{enumitem}
\usepackage{cite}
\usepackage{booktabs}
\usepackage{amsmath,amssymb,amsfonts}
\usepackage{algorithmic}
\usepackage{setspace}
\usepackage{subcaption} 
\usepackage{graphicx}
\usepackage{textcomp}
\usepackage[table,xcdraw]{xcolor}
\usepackage{graphicx} 
\usepackage{multirow}
\usepackage{threeparttable}
\usepackage{tabularx}
\usepackage{mathtools}
\usepackage{caption}
\usepackage{pifont}
\usepackage{nameref}

\newcommand{\bluetext}[1]{\textcolor{black}{#1}}
\newcommand{\bluebf}[1]{\textcolor{black}{\textbf{#1}}}

\definecolor{gradient1}{RGB}{187, 210, 160} 
\definecolor{gradient2}{RGB}{227, 230, 189}
\definecolor{gradient3}{RGB}{253, 240, 207} 
\definecolor{gradient4}{RGB}{254, 232, 204} 
\definecolor{gradient5}{RGB}{246, 206, 190}
\definecolor{gradient6}{RGB}{231, 178, 172}

\definecolor{g9}{RGB}{238, 192, 181}
\definecolor{g8}{RGB}{246, 206, 190}
\definecolor{g7}{RGB}{252, 220, 198}
\definecolor{g6}{RGB}{254, 232, 204}
\definecolor{g5}{RGB}{254, 236, 205}
\definecolor{g4}{RGB}{253, 240, 207}
\definecolor{g3}{RGB}{227, 230, 189}
\definecolor{g2}{RGB}{187, 210, 160}
\definecolor{g1}{RGB}{168, 201, 146}

\usepackage[free-standing-units=true]{siunitx}
\newcommand{\rom}[1]{\uppercase\expandafter{\romannumeral #1\relax}}
\usepackage{todonotes}
\usepackage{array}
\newcommand{\PreserveBackslash}[1]{\let\temp=\\#1\let\\=\temp}
\newcolumntype{C}[1]{>{\PreserveBackslash\centering}p{#1}}
\newcolumntype{R}[1]{>{\PreserveBackslash\raggedleft}p{#1}}
\newcolumntype{L}[1]{>{\PreserveBackslash\raggedright}p{#1}}

\newcommand{\fref}[1]{Fig. \ref{#1}}
\newcommand{\sref}[1]{Section. \ref{#1}}
\newcommand{\tref}[1]{Table. \ref{#1}}

\newcommand{\aref}[1]{Appendix. \ref{#1}}
\newcommand{\iref}[2]{\ref{#1} #2}

\makeatletter
\let\NAT@parse\undefined
\makeatother

\usepackage[colorlinks]{hyperref}
  \hypersetup{
    citecolor=blue,
    linkcolor=blue,   
    urlcolor=blue}

\begin{document}

% \title{\LARGE \bf
% AirIO: Learning Inertial Odometry with Enhanced IMU feature Observability\\
% \bf\Large{\href{https://air-io.github.io}{air-io.github.io}}
% }

\title{AirIO: Learning Inertial Odometry with Enhanced \\IMU Feature Observability\\{\fontsize{15pt}{5pt}\selectfont\href{https://air-io.github.io}{air-io.github.io}}}
% \title{\LARGE \bf
% AirIO: Learning Inertial Odometry with Enhanced \\IMU Feature Observability\\
% \bf\Large{\href{https://air-io.github.io}{air-io.github.io}}
% }

\author{Yuheng Qiu$^{*}$$^{1}$, Can Xu$^{*}$$^{1}$, Yutian Chen$^{1}$, Shibo Zhao$^{1}$, Junyi Geng$^{2}$ and Sebastian Scherer$^{1}$% <-this % stops a space
\thanks{Manuscript received: February, 15, 2025; Revised April, 30, 2025; Accepted June, 1, 2025.}%Use only for final RAL version
\thanks{This paper was recommended for publication by Editor Giuseppe Loianno upon evaluation of the Associate Editor and Reviewers' comments.} %Use only for final RAL version

\thanks{$^{*}$Equal contribution.}
\thanks{$^{1}$Yuheng Qiu, Can Xu, Yutian Chen, Shibo Zhao, and Sebastian Scherer are with the Robotics Institute, Carnegie Mellon University, Pittsburgh, PA 15213, USA {\tt\small \{yuhengq, canxu, yutianch, shiboz, basti\} @andrew.cmu.edu;} $^{2}$Junyi Geng is with Department of Aerospace Engineering, Pennsylvania State University, University Park, PA, 16802, USA {\tt\small jgeng@psu.edu}.}%
\thanks{Digital Object Identifier (DOI): see top of this page.}
}

\thispagestyle{empty}
\pagestyle{headings}
% \maketitle
\makeatletter
\g@addto@macro\@maketitle{
  \captionsetup{type=figure}\setcounter{figure}{0}
  \def\mycolspace{1.2mm}
  \centering
\includegraphics[width=2\columnwidth]{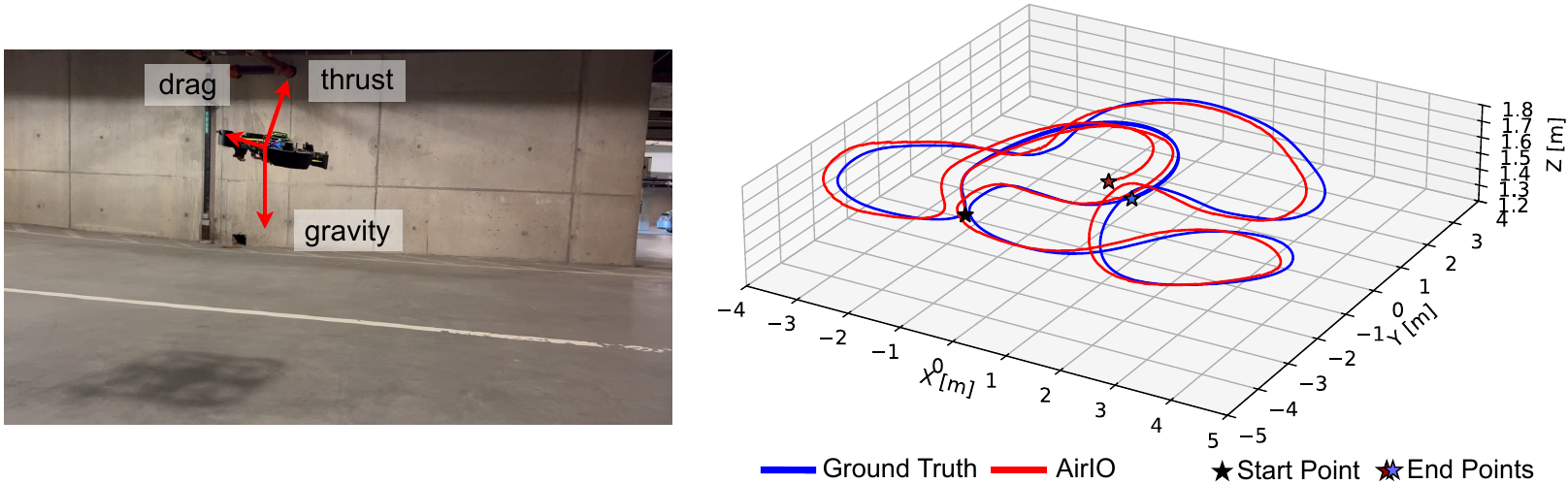}
	\captionof{figure}{When deploying learning-based IO on drones, we observed that preserving the IMU feature representation in the body frame while retaining gravitational acceleration improves the ATE by \textbf{73.9\%} on the Blackbird dataset.
    This significant improvement occurs because preserving the IMU data in the body frame, along with gravitational information, enhances observability and retains more kinematic details.
    Based on this finding, we propose AirIO, 
    which outperforms state-of-the-art algorithms without extra information like control signals or additional sensors.}
	\label{fig:eyecatcher}
}

\makeatother

\maketitle
\begin{abstract}
Inertial odometry (IO) using only Inertial Measurement Units (IMUs) offers a lightweight and cost-effective solution for Unmanned Aerial Vehicle (UAV) applications, yet existing learning-based IO models often fail to generalize to UAVs due to the highly dynamic and non-linear-flight patterns that differ from pedestrian motion. 
In this work, we identify that the conventional practice of transforming raw IMU data to global coordinates undermines the observability of critical information in UAVs. By preserving the body-frame representation, our method achieves substantial performance improvements, with a 66.7\% average increase in accuracy across three datasets. Furthermore, explicitly encoding attitude information into the motion network results in an additional 23.8\% improvement over prior results. Combined with a data-driven IMU correction model (AirIMU) and an uncertainty-aware Extended Kalman Filter (EKF), our approach ensures robust state estimation under aggressive UAV maneuvers without relying on external sensors or control inputs. Notably, our method also demonstrates strong generalizability to unseen data, underscoring its potential for real-world UAV applications.
\end{abstract}

\begin{IEEEkeywords}
Aerial Systems: Perception and Autonomy; Deep Learning Methods; Localization
\end{IEEEkeywords}

\section{Introduction}

\IEEEPARstart{I}{nertial} Measurements Units (IMUs) are inexpensive and ubiquitous sensors that provide linear accelerations and angular velocities.  
Due to the size, weight, power, and cost (SWAP-C) constraints, IO based only on light and low-cost sensors is ideal for Unmanned Aerial Vehicles (UAVs) applications ranging from 3D mapping \cite{zhao2021super}, exploration \cite{hu2023off} to physical interaction~\cite{guo2024flying}.
% and package delivery~\cite{geng2021estimation}.  
Compared to exteroceptive sensors like vision and LiDAR, IMUs are unaffected by visual degradation factors such as motion blur~\cite{Zhao_2024_CVPR} or dynamic object interruption~\cite{qiu2022airdos}, which are common in agile UAV flights~\cite{CioffiRal2023}.
Despite these advantages, current IO solutions struggle to accurately adapt to the complex motion models of UAVs due to the IMU inherent noise and bias, leading to large drift over time, particularly during agile maneuvers.

Most recent advances in learning-based IO have focused on pedestrian and legged robots \cite{liu2020tlio, herath2020ronin, buchanan2022deep}, utilizing motion priors like stride length~\cite{yan2018ridi} and repetitive gait patterns~\cite{yang2023multi, teng2020arpdr} beyond IMU pre-integration~\cite{forster2015imu}. 
In contrast, multirotor UAV motion involves rigorous maneuvers affecting orientation and thrust, resulting in highly dynamic, nonlinear behavior without clear priors. 
Small attitude changes can cause significant variations in velocity and position, complicating the IO process and making pedestrian-focused methods less effective~\cite{CioffiRal2023}. 
As a result, deploying learning-based IO on UAVs often requires additional sensors, such as tachometers~\cite{zhang2022dido} or control inputs like thrust commands~\cite{CioffiRal2023}.
This raises our main questions: 

\textit{Why is learning IO not directly applicable to UAVs?\\How can learning IO be effectively deployed for UAVs?}
% \begin{figure}[t]
%     \centering
%     \includegraphics[width=1\linewidth]{figs/airio_v.png}
%     \caption{
%     AirIO achieves state-of-the-art performance by relying only on information from IMU.
%     We find out that simply changing the IMU feature representation can improve 74\% of the ATE on the BlackBird dataset.}
%     \label{fig:enter-label}
% \end{figure}
% \begin{figure*}[t]
%     \centering
%     \includegraphics[width=1\linewidth]{figs/airio_h.drawio.png}
%     \caption{
%     AirIO achieves state-of-the-art performance by relying only on information from IMU.
%     We find out that simply changing the IMU feature representation can improve 74\% of the ATE on the BlackBird dataset.}
%     \label{fig:enter-label}
% \end{figure*}

% However, in practice, deploying IO on UAV is challenging due to the inherent noise and bias of the IMU measurement, which is even significant under dynamic and unpredictable environments that are typical for quadrotors, causing large drift over time. 
% In some dynamic scenarios, such as helicopter cruising, learning-based IO has even been shown to underperform compared to basic double integration with a well-calibrated IMU \cite{qiu2023airimu}.
In this paper, we investigate the effective representation of learning-based IO in multirotor UAV motion and propose a solution that relies solely on IMU data. Through rigorous feature analysis and examination of UAV dynamics, we identify that the commonly used global-frame representation is less effective for dynamic, agile maneuvers due to the less observable attitude information compared to IMU data in the body-frame representation. To address this, we introduce AirIO, a learning-based IO method that explicitly encodes attitude information and predicts velocity using body-frame representation. We further integrate an EKF that fuses IMU pre-integration with the learning-based model for odometry estimation.
We show that our approach outperforms state-of-the-art algorithms \cite{CioffiRal2023} in various scenarios and environments, even without additional sensors or control information. %Notably, simply changing the IMU body representation contributes to an 82.6\% improvement in the drone racing datasets. 

The main contributions of this work are: \begin{enumerate} 
% \item We conduct the analysis of the extracted features from different representations using t-SNE and condition numbers of the principal components. The results show that the body frame representation is more effective and observable.
% \item Based on the effective representation, we develop a learning-based IO framework AirIO to explicitly encode the attitude information and fuse it with the raw IMU data to predict the velocity. We further integrate an uncertainty-aware IMU preintegration model and the learned model into EKF for odometry estimation. 
% \item We validate our approach on multiple real-world experiment data and compare it with the state-of-the-art IO algorithms. Experiments show that simply preserving the input representation can significantly improve IO accuracy by 74.4\%. Encoding the attitude information into the framework further improves the estimation accuracy by 78.5\%. 
\item We conduct the analysis of the extracted features from different representations using \bluetext{t-Distributed Stochastic Neighbor Embedding (t-SNE)} and principal component analysis (PCA). The results show that body-frame representation is more expressive and observable.
Simply changing the input representation can significantly improve IO accuracy by an average of 66.7\%.
\item Building on this, we develop AirIO, a learning-based IO that encodes the attitude information and fuses it with the body-frame IMU data for velocity prediction. This explicit encoding further improves accuracy by 23.8\%.
Additionally, we integrate an uncertainty-aware IMU preintegration model and a learned motion network into EKF for odometry estimation.
\item  We validate our approach on extensive experiments. It outperforms existing IO algorithms without the need for additional sensors or control information. It also demonstrates generalizability to the unseen datasets.
\end{enumerate}

\section{Related Work}
\subsection{Model-based Inertial Odometry}
Traditional model-based inertial odometry (IO) methods rely on kinematic models \cite{forster2015imu} to estimate relative motion and are often integrated with exteroceptive sensors such as cameras \cite{qin2018vins} and LiDAR \cite{zhao2021super}. While these methods achieve high accuracy, their dependence on external sensors makes them vulnerable to disturbances caused by agile motion or dynamic environments.
To address the SWAP challenge of lightweight UAV navigation, Ref~\cite{svacha2020imu} introduced a method that estimates tilt and velocity by fusing tachometer and IMU data. 
Although effective, this approach still relies on auxiliary sensors, limiting its applicability in environments where such sensors are unavailable or unreliable.

\subsection{Learning Inertial Odometry}
Recent advances in deep learning have sparked research in data-driven methods for learning velocity and displacement from inertial data \cite{yan2018ridi, chen2018ionet, herath2020ronin, herath2022neural}. 
Approaches such as IONet \cite{chen2018ionet}, A2DIO \cite{wang2022a2dio}, and NILoc \cite{herath2022neural} formulate inertial navigation as a sequential task by predicting relative position displacements using IMU data. 
Subsequent work has incorporated these learned displacements into EKF \cite{liu2020tlio, sun2021idol, deng2023data} or batched optimization frameworks \cite{buchanan2022learning}.
To ensure consistency in IMU measurements, many of these methods transform the input IMU data into global coordinates \bluetext{using ground-truth orientation}.
For pedestrian navigation, RoNIN \cite{herath2020ronin} proposed a \bluetext{Heading-Agnostic Coordinate Frame} (HACF), aligning the accelerometer’s gravity vector with the z-axis and constraining rotation to the horizontal plane. This framework has been widely adopted for learning-based IO \cite{liu2020tlio, sun2021idol} due to its effectiveness in simplifying pedestrian motion. 
Building on HACF, RIO \cite{cao2022rio} introduced rotation-equivalence data augmentation to self-supervise pose estimation after orientation alignment. 
\bluetext{Since pedestrians and wheeled vehicles predominantly move along the horizontal plane, HACF effectively removes yaw dependency from the representation, thereby reducing learning complexity and improving generalizability.}
% Due to this simplicity, most of the existing works also use the HACF or global coordinate frame on other modalities.
Despite the success in pedestrian navigation and wheeled robots \cite{herath2020ronin, brossard2019rins}, these representations are less effective for multirotor UAV motion, which requires more sophisticated dynamic modeling.
In this paper, we show that the commonly used global coordinate representations hinder neural networks from effectively capturing the dynamic complexities of UAVs.

\subsection{Inertial Odometry for Multirotor UAVs}

% It is challenging to deploy inertial odometry on the quadrotor due to the rigorous movement. 
 
% To address this limitation, 
Existing learning-based IO methods have primarily focused on pedestrian datasets, limiting their generalizability to platforms with more demanding motion dynamics, such as UAVs.
Approaches like AI-driven pre-processing and down-sampling of high-speed inertial data have been proposed for better state estimation~\cite{steinbrener2022improved}.
\bluetext{Due to its simplicity, most of the existing methods continue adopting the global coordinate frame, such as~\cite{zhang2022dido, CioffiRal2023, bajwa2024dive}}.
DIDO~\cite{zhang2022dido} estimates the thrust of UAVs using tachometer data and addresses unmodeled forces by incorporating a neural network.
Building on this, the IMO~\cite{CioffiRal2023} incorporates thrust information and IMU data within an EKF framework, which implicitly captures the drone dynamics. 
However, IMO relies on additional control signals to capture the drone's dynamics, as shown in \fref{fig:unseen_sid}, which may suffer from overfitting on training datasets.
DIVE \cite{bajwa2024dive}, a more recent work for UAV IO, encodes orientation alongside gravity-removed global-frame acceleration. 
While this representation improves the robustness of learning-based IO, our findings in \sref{Sec: Ablation} indicate that it remains suboptimal for capturing UAV dynamics comprehensively.
% In this paper, we present an algorithm that surpasses existing methods in performance without additional information, offering a more robust and generalizable solution.
\section{Methodology}

%The goal of learning IO is to estimate the relative position transform (velocity) $\mathbf{v}$ from the IMU data $\{\mathbf a_i, \mathbf w_i \}$ which is defined as $\mathbf v = \Psi(\{\mathbf a_i, \mathbf w_i\})^n_{i=1} \in \mathbb{R}^3$ from the local body IMU frame.
The goal of learning IO is to estimate relative position transform (velocity) from the IMU data in the local body IMU frame.
Most existing methods employ transforming the IMU frame to global frame or removing gravity from the raw IMU data\cite{herath2020ronin, cao2022rio,bajwa2024dive}. 
% Existing methods pre-process IMUgravity removal \cite{herath2020ronin, cao2022rio,bajwa2024dive}.  measurements through techniques such as gravity alignment, and gravity removal \cite{herath2020ronin, cao2022rio,bajwa2024dive}. 
% As illustrated in the \fref{fig:network-struct}, we propose AirIO system that consists of three modules: feature selection, 
Our study reveals that these widely used representations, such as global coordinate frames, HACF, and gravity-removed frames, limit the effectiveness of IMU feature extraction. Thus, this section proposes a different feature representation and the corresponding state estimation design. It consists of three parts: feature representation analysis, attitude-encoded network design, and a tightly coupled EKF system.

%We first investigate the feature representation of the learning IO in \sref{sec:coordinate} and \sref{sec:featurerepresentation}, finding that the body coordinate frame is more representative than the other representation. 
%Grounded on these findings, we propose AirIO which explicitly encode the drone's pose to predict the body-frame velocity ${}^\mathcal{B}\hat{\mathbf{v}}_i$ and the corresponding covariance $\hat{\boldsymbol{\Sigma}}^v_i$ in \sref{sec:network}. 
%Lastly, in \sref{sec:EKF} we build a tightly coupled EKF system that fuses the AirIO motion network with the learning-based IMU preintegration network, AirIMU \cite{qiu2023airimu}, which jointly estimates the uncertainty of the IMU preintegration. Both modules predict their corresponding uncertainties, contributing to improved overall performance and robustness.

% Grounded in our previous analysis of IMU feature representation, we conclude that body-frame features are more representative, and attitude information is crucial for accurate drone modeling.

\subsection{IMU Coordinate Frame}
\label{sec:coordinate}
\textbf{Canonical Body Coordinate Frame:} 
The IMU accelerometer measures the net force per unit mass acting on the sensor.
Due to the measurement mechanism, it measures not only the object actual accelerations due to motion ${}^\mathcal{B}\mathbf{\dot v}= \frac{{}^\mathcal{B}\mathbf{F}_{{net}_i}}{m}$ in the sensor's local coordinate frame (usually aligned with object body frame), but also the constant acceleration due to Earth's gravity ${}^\mathcal{G}\mathbf{g}$ directed toward the Earth’s center.
%including the gravitational acceleration $\mathbf{g}^G$ is directed toward the Earth’s center, and the net acceleration applied to the sensor in the body frame, $ \mathbf{\dot v}^B= \frac{\mathbf{F}^B_{{net}_i}}{\mathbf{m}}$. 
Thus, the accelerometer measurement at timestamp $i$ as:
% As illustrated in \fref{fig:attitude}, the accelerometers measurement is defined by:
\begin{equation}
{}^\mathcal{B}\mathbf{a}_i = \frac{{}^\mathcal{B}\mathbf{F}_{{net}_i}}{m} +  \mathbf{R}_i^\top {}^\mathcal{G}\mathbf{g},
\label{eq:body}
\end{equation}
where $m$ is the object mass, and $\mathbf{R}_i \in \mathbb{SO}(3)$ is the rotation for vector from body frame $\mathcal{B}$ to the global frame $\mathcal{G}$. ${}^\mathcal{F}(\cdot)$ denotes quantity represented in frame $\mathcal{F}$. 
% Detailed UAV dynamics can be found in \aref{Sec:QuadrotorModel}.

% Of these components, gravitational acceleration is the dominant force consistently directed toward the Earth’s center in the global coordinate frame $\mathbf{g}^G = [0, 0, 9.81]^T$. 
% In \eref{eq:body}, the gravity term $\mathbf{R_i}^T \mathbf{g}$ combines gravitational acceleration with the drone’s attitude $\mathbf{R_i}$.
% Dominated by gravitational acceleration, the drone’s attitude information is more observable in the body-frame IMU feature representation.
\begin{figure}
    \centering
    \includegraphics[width=1\linewidth]{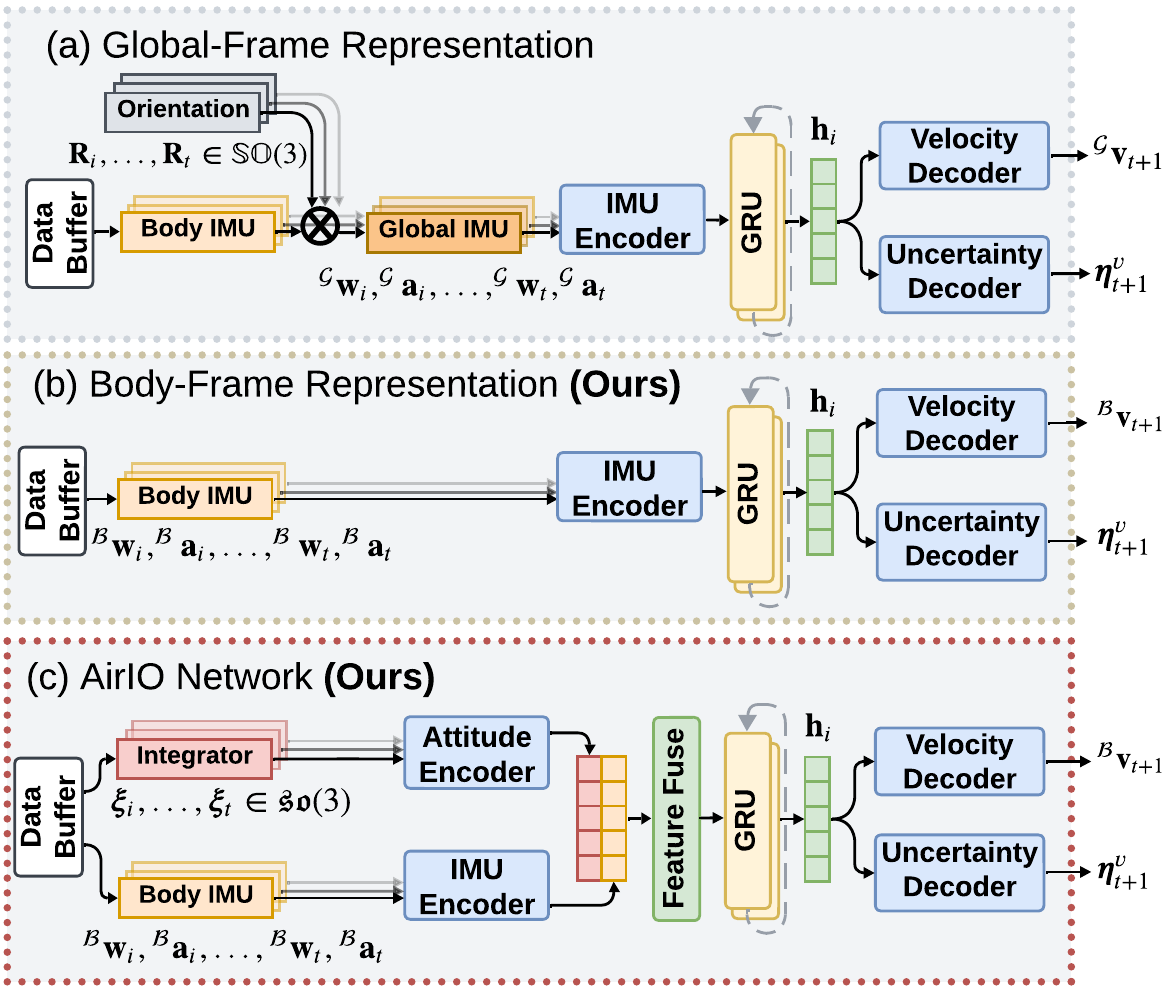}
    \caption{(\textbf{a}) Existing learning IO usually transforms the IMU input to the global coordinate using ground-truth orientation. (\textbf{b}) We found that preserving the IMU data in the body coordinate, along with gravitational information, can significantly improve accuracy.
    (\textbf{c}) Our AirIO model leverages the body-frame IMU and explicitly encodes the UAV's orientation to estimate the body-frame velocity.
    % We utilize one CNN encoder to capture IMU features from the accelerometer and gyroscope, and one encoder for the drone's orientation.
    % The output features are concatenated and fused via an MLP. 
    % GRU then encodes the fused feature to capture temporal information. 
    % We use the encoded features to predict the body coordinate frame velocity and its covariance. 
    }
    \vspace{-10pt}
    \label{fig:network}
\end{figure}

\textbf{Global Coordinate Frame:}
Many existing methods transform the IMU data from body frame $\mathcal{B}$ to HACF or global frame $\mathcal{G}$ by applying an estimated transform rotation $\hat{\mathbf{R}}_i$, e.g. 
%by aligning the gravitational acceleration to the z-axis of the global coordinate frame. 
%In general, these methods apply a predicted transform rotation $\mathbf{\hat R_i}$, which transforms the IMU from frame $\{\mathcal{B}\}$ to frame $\{\mathcal{G}\}$. 
\begin{equation}
{}^\mathcal{G}\mathbf{a}_i = \hat{\mathbf{R}}_i \frac{{}^\mathcal{B}\mathbf{F}_{{net}_i}}{m} +  \hat{\mathbf{R}}_i\mathbf{R}_i^\top{}^\mathcal{G}\mathbf{g}
\label{eq:global}
\end{equation}
The rotation $\hat{\mathbf{R}}_i$ can often be obtained by simple IMU preintegration or the estimation results from an EKF, which is often accurate enough, such that $\hat{\mathbf{R}}_i \mathbf{R}_i^\top \approx \mathbf{I}$.
%If the estimation is accurate enough, the estimated $\mathbf{\hat R_i}$ can approximate $\mathbf{\hat R_i} \mathbf{R_i}^T \approx \mathbf{I}$.s

Notice that both (\ref{eq:body}) and (\ref{eq:global}) represents the same physical process. The key difference lies in how the attitude information is coupled with different components: in (\ref{eq:body}), it is coupled with the gravitational force, whereas in (\ref{eq:global}), it is coupled with the motion force. In (\ref{eq:body}), since the gravity acceleration ${}^\mathcal{G}\mathbf{g}$ is a constant, the the motion force and attitude form a linear combination. In contrast. in (\ref{eq:global}), the attitude couples with motion force in a nonlinear manner, with gravity acting as an additional constant that does not contribute extra useful information. Thus, the motion force and attitude are intuitively easier to estimate in (\ref{eq:body}), while (\ref{eq:global}) requires a more complex network structure for estimation.

\subsection{Representation Analysis}
\label{sec:featurerepresentation}
We perform feature analysis to verify our intuition. Specifically, we extract the latent features $\mathbf h_i$ from the IMU feature encoder for the input under different representations in \fref{fig:network}. 
%This latent feature, denoted as $h_i$ in \fref{fig:network-struct} serves as the core representation for the learning framework. 
We conduct multiple feature analyses to assess their effectiveness and representativeness.

\textbf{Principal Component Analysis} PCA is a statistical method that reduces the dimensionality of a data set by replacing correlated variables with a smaller set of uncorrelated principal components. To analyze the expressivity of the input representation, we performed PCA on the feature latent space for different input coordinates. 
% Then, we perform PCA on the latent feature $M$ to analyze its underlying structure by computing the singular value of the normalized feature matrix in a numerical stable way. Each singular value represents the corresponding magnitude scaling of one principle feature component. In particular, we quantify the energy coverage (or the explained variance) for the top $k$ principal components in PCA. It shows that the top $k = 10$ principle components already covers 95\% of the total energy, which can be used to represent the latent feature space. 
We concatenate the latent features from all samples to form a feature matrix and perform PCA on the latent feature matrix to analyze its underlying structure by computing the singular value of the normalized feature matrix in a numerically stable way. Each singular value represents the corresponding magnitude scaling of one principal feature component. We then quantify the cumulative explained variance (or the energy coverage) for the top $k$ principal components in PCA. 
Specifically, we evaluate on two datasets: the widely used UAV dataset Blackbird\cite{antonini2018blackbird} collected for agile autonomous flight, which covers large UAV maneuvers, and our custom simulation dataset Pegasus, which which was collected in the NVIDIA IssacSim simulator with Pegasus autopilot \cite{Jacinto2024pegasus}, featuring diverse maneuvers in an ideal environment without external disturbance, allowing us to work with cleaner data.
%Additionally, we apply the same process to our custom simulation dataset Pegasus dataset, which we selected using the Pegasus simulator\cite{Jacinto2024pegasus}. Utilizing simulation data allows us to work with cleaner data by eliminating external disturbances. This approach provides a more precise foundation for our feature analyses.
\begin{figure}[H]
    \centering
    \includegraphics[width=1\linewidth]{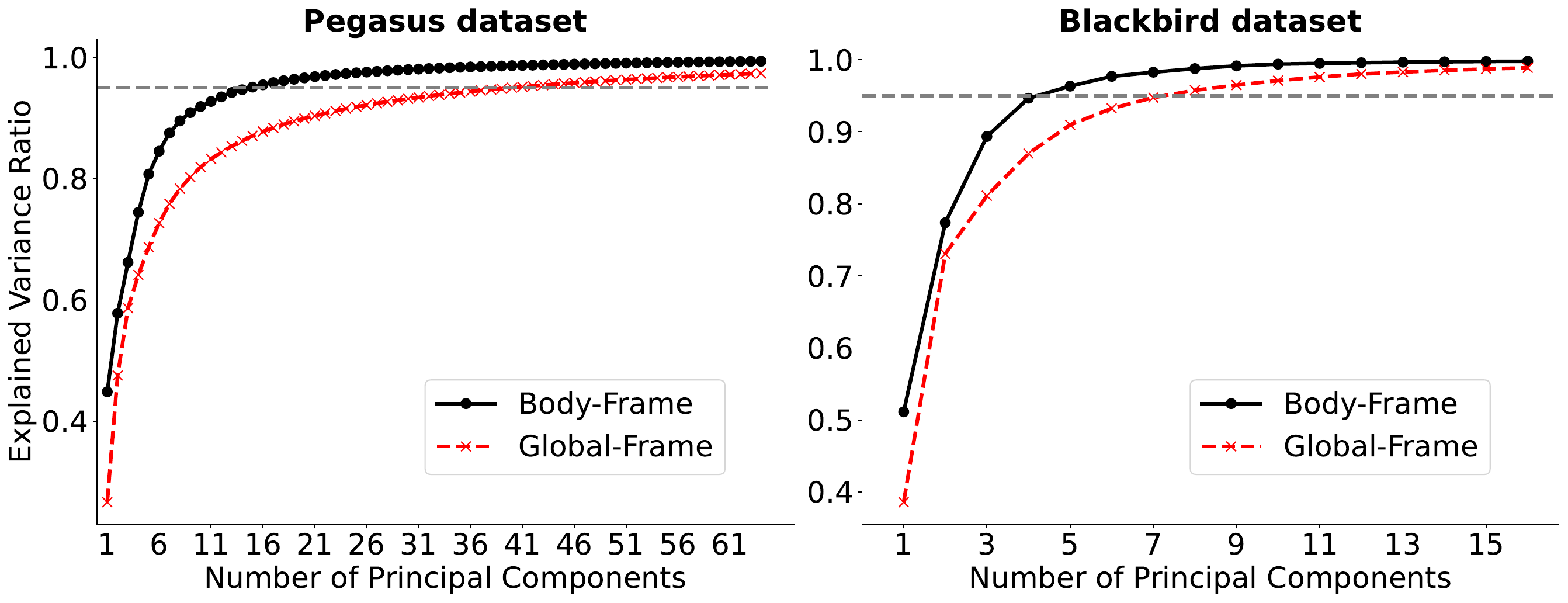}
    \caption{The cumulative explained variance of the latent features for the Pegasus dataset (\textbf{Left}) and Blackbird dataset (\textbf{Right}). In both datasets, the latent features derived from body frames (black line) require fewer principal components to achieve comparable total energy.}
    \label{fig:svd}
\end{figure}

\begin{figure}[!t]
    \centering
    \includegraphics[width=1\linewidth]{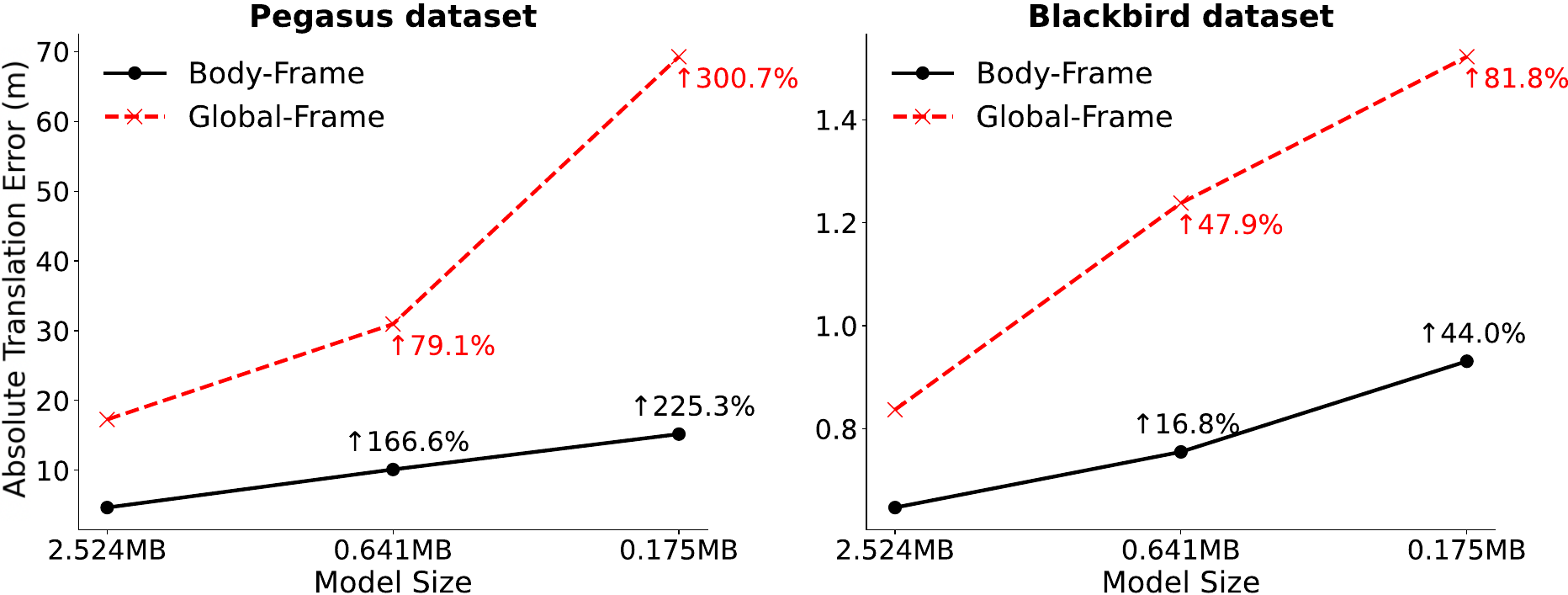}
    \caption{The ATE for body-frame and global-frame models at different model sizes on Pegasus dataset (\textbf{Left}) and Blackbird dataset (\textbf{Right}). The text denotes the relative percentage increase in ATE. }
    \vspace{-10pt}
    \label{fig:model_ablation}
\end{figure}

\fref{fig:svd} shows the cumulative energy coverage of the latent feature derived from two different IMU input representations: global and body frames. The red and black curves represent the energy coverage of the top $k$ principal components. Specifically, for the Blackbird dataset, it shows that the top $k$ = 5 principal components in body-frame representation already cover 95\% of the total energy. In contrast, global-frame representation requires the top $k$ = 8 principal components to achieve the same level of energy. 
%The largest explained variance from the body frame is about 1.7 times that of the global frame. 
Similarly, in the Pegasus datasets, the body-frame representation requires top $k$ = 15 principal components and the global frame require top $k$ = 40 principal components to cover 95\% of the total energy. It clearly shows that models trained on the body representation input can achieve comparable performance with fewer features, highlighting the efficiency and expressivity of this representation in capturing the essential features. It also implies that models trained under body frame have better compressibility and can be designed more lightweight.
\fref{fig:model_ablation} shows the comparison of the absolute translation error (ATE) for the models at different scales under both body and global frames. As the model size decreases, the prediction performance for model under body-frame representation degrades more smoothly and consistently yields lower errors. See \aref{Appendix:Model Size} for more studies on the model compressibility.

\begin{figure}[H]
    \centering
    \includegraphics[width=1\linewidth]{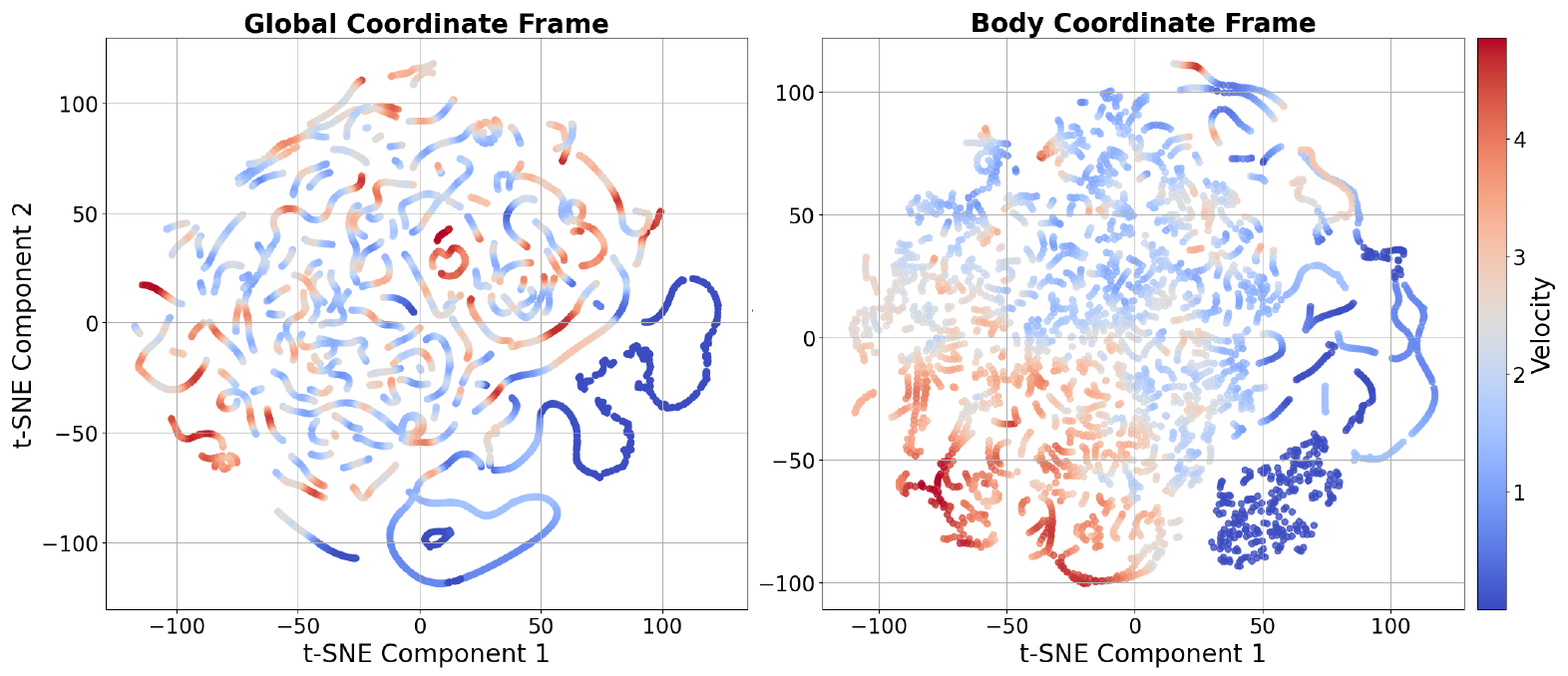}
    \caption{t-SNE analysis on Pegasus dataset with each point colored based on the velocity magnitude from low to high, represented by blue to red.} % \textbf{Left:} The latent features are highly entangled in the feature obtained from the global coordinate frame. \textbf{Right:} The latent features from the body coordinate frame are well-separated and exhibit clearer structure corresponding to different velocities.}
    \label{fig:tsne}
\end{figure}

\textbf{t-SNE}~\cite{van2008visualizing} We also perform t-SNE on Pegasus dataset, a dimensionality reduction approach that projects high-dimension data into a low-dimensional map to analyze the feature representation qualitatively.
% In the Pegasus dataset, we utilized t-SNE to project the high-dimensional latent feature distribution in a 2D space. 
As shown in \fref{fig:tsne}, based on the network prediction (here is the velocity), we color code each data point by the velocity magnitude from low to high. In the global frame, the points exhibit a highly entangled distribution and different velocity levels overlap, indicating poor separability and less representative encoding. 
In contrast, the feature distribution of the body coordinate frame is well-separated, where different velocity levels form more coherent clusters. This suggests that the body frame facilitates a more discriminative and effective representation of latent features.%, likely due to its alignment with the underlying dynamics of the IMU data.

% Overall we can draw two conclusions from our analysis:
% % In global-frame IMU representation, the kinematic information (thrust $F$) is conjugated with the rotation $R$, creating a nonlinear coupling that obscures the neural network's ability to effectively capture the kinematic information. This observation is further supported by the t-SNE analysis. 
% \begin{itemize}
%     \item The nonlinear coupling between thrust F and rotation R in global-frame IMU representation limits the neural network's ability to capture kinematic information, as supported by t-SNE analysis.
%     \item The body-frame representation proves to be more expressive and discriminative, as demonstrated by PCA analysis and the compressibility experiment.
% \end{itemize}

In summary, the analysis from \sref{sec:coordinate} and \sref{sec:featurerepresentation} indicates that \romannumeral 1). Good prior information will help for the estimation since it couples with the motion force in a linear or nonlinear manner, as shown in (\ref{eq:body})(\ref{eq:global}).
\romannumeral 2). The nonlinear coupling between the motion force and attitude in the global-frame representation complicates the estimation problem. Body-frame representation proves to be more expressive and discriminative, leading to a light network for prediction.
%\update{We believe that the proposed body-frame representation can be adopted by the general robots. However, UAVs tend to benefit more from it because of their agile motion characteristics. Unlike many ground robots that primarily operate with yaw-dominant movement, UAVs often exhibit significant roll and pitch maneuvers. As a result, the linear coupling between motion force and attitude inherent in the body-frame representation enables it to more effectively capture the dynamics of UAVs and achieve accurate motion estimation.}
{Notice that while the analysis is indeed applicable to general robots via standard IMU equations, UAVs present a compelling test case due to their agile motion. Unlike many mobile or legged robots with mainly yaw-dominant movement, UAVs often exhibit significant roll and pitch maneuvers, resulting in more pronounced impact of frame representation difference.}

\subsection{AirIO Motion Network \& Training}
\label{sec:network}
% However, during periods of significant UAV acceleration, gravitational acceleration becomes less dominant, reducing the observability of kinematic information. 
% To overcome this limitation, we explicitly encode the drone's attitude within the motion network.
Based on the previous analysis, 
%we observed that the attitude information is inherently entangled with other kinematic information in the IMU data, making it challenging for the neural network to disentangle them explicitly. To address this limitation, 
we introduce a separate encoding of the drone's attitude $\boldsymbol{\xi} \in {\mathfrak{so}(3)}$ to the motion network beyond the IMU encoder with body-frame representation, shown in \fref{fig:network}(c). After fusing this information, the network predicts the body-frame velocity ${}^\mathcal{B}\hat{\mathbf{v}}$ and the corresponding uncertainty $\hat{\boldsymbol{\eta}}^v$.
The motion network can be written as: $({}^\mathcal{B}\hat{\mathbf{v}}, \hat{\boldsymbol{\eta}}^v) = f_{\boldsymbol{\theta}}({}^\mathcal{B}\mathbf{w}, {}^\mathcal{B}\mathbf{a}, \boldsymbol{\xi})$.

Inspired by \cite{qiu2023airimu}, we map the drone's orientation to the ${\mathfrak{so}(3)}$ Lie algebra space for compact and continuous representation of 3D rotations and smoother network gradient to mitigate numerical instability. We then leverage a CNN encoder to encode ${\mathfrak{so}(3)}$. 
%To effectively integrate attitude information with the IMU data encoder, we map the drone's orientation to the ${\mathfrak{so}(3)}$ space before feeding it into the neural network.
%Using Lie algebra to encode orientation provides a compact, unconstrained, and continuous representation of 3D rotations that naturally aligns with rotational geometry. 
%This approach not only mitigates numerical instability but also promotes smoother gradient flow during training, resulting in improved network performance.

% Our network leverage parallel CNN-based encoders to separately encode the drone's attitude information $\boldsymbol{\xi} \in {\mathfrak{so}(3)}$ and body-frame IMU data $(\mathbf{w}^\mathcal{B}, \mathbf{a}^\mathcal{B})$.  
% We use CNN encoders to extract the features from the body-frame IMU data $(\mathbf{w}^\mathcal{B}, \mathbf{a}^\mathcal{B})$ and the orientation of the drone $\boldsymbol{\xi} \in {\mathfrak{so}(3)}$.
After concatenating the features from the attitude encoder and the IMU encoder, we employ bi-directional GRU layers to extract the latent feature $\mathbf{h}_i$ and model the temporal dependencies in the drone dynamics. We leverage this feature to estimate the body-frame velocity and the corresponding uncertainty by two MLP decoders.

\bluetext{To jointly supervise the velocity and the corresponding uncertainty ${}^\mathcal{B}\hat{\mathbf{v}}_i$, we designed a combined loss function: 
\begin{equation}
   \mathcal{L} = \mathcal{L}_{\text{Huber}} + \lambda \mathcal{L}_{C},
\end{equation}}
\bluetext{where the $\lambda = 1e^{-4}$. The $\mathcal{L}_{\text{Huber}}$ is the Huber loss between the predicted body-frame velocity ${}^\mathcal{B}\mathbf{v}_i$ and the body-frame ground-truth velocity ${}^\mathcal{B}\hat{\mathbf{v}}_i$, defined as:}
\bluetext{
\begin{equation}
    \mathcal{L}_{\text{Huber}} = \begin{cases}
        \frac{1}{2}({}^\mathcal{B}\mathbf{v}_i - {}^\mathcal{B}\hat{\mathbf{v}}_i)^2 & \text{if } |{}^\mathcal{B}\mathbf{v}_i - {}^\mathcal{B}\hat{\mathbf{v}}_i| < \delta \\
        \delta \cdot \left( |{}^\mathcal{B}\mathbf{v}_i - {}^\mathcal{B}\hat{\mathbf{v}}_i| - \frac{1}{2}\delta \right) & \text{otherwise}
    \end{cases} ,
\end{equation}
where we set the $\delta = 0.005$.}

To supervise the uncertainty of the velocity, we employ an uncertainty loss inspired by~\cite{liu2020tlio}, assuming that the estimated velocity uncertainty follows a Gaussian distribution:
\begin{equation}
    \bluetext{\mathcal{L}_{C}} = ({}^\mathcal{B}\mathbf{v}_i - {}^\mathcal{B}\hat{\mathbf{v}}_i) \hat{\boldsymbol{\Sigma}}^{v}_{i}{}^{-1} {({}^\mathcal{B}\mathbf{v}_i - {}^\mathcal{B}\hat{\mathbf{v}}_i})^T +\ln (\det \hat{\boldsymbol{\Sigma}}^{v}_i),
\end{equation}
where $\hat{\boldsymbol{\Sigma}}^{v}_i$ is $3 \times 3$ covariance matrix for the $i^{th}$ data. Similar to the setting in \cite{liu2020tlio}, we simplified the covariance as  $\hat{\boldsymbol{\Sigma}}^{v}_i = \text{diag} (\hat{\boldsymbol{\eta}} ^{v}_i{}^2)$, where $\hat{\boldsymbol{\eta}} ^{v}_i$ is learned uncertainty from network.

For orientation $\boldsymbol{\xi}$, we use the ground truth during training and replace it with EKF-estimated values during testing.

\subsection{Extended Kalman Filter}
\label{sec:EKF}

\begin{figure}[!t]
    \centering
    \includegraphics[width=1\linewidth]{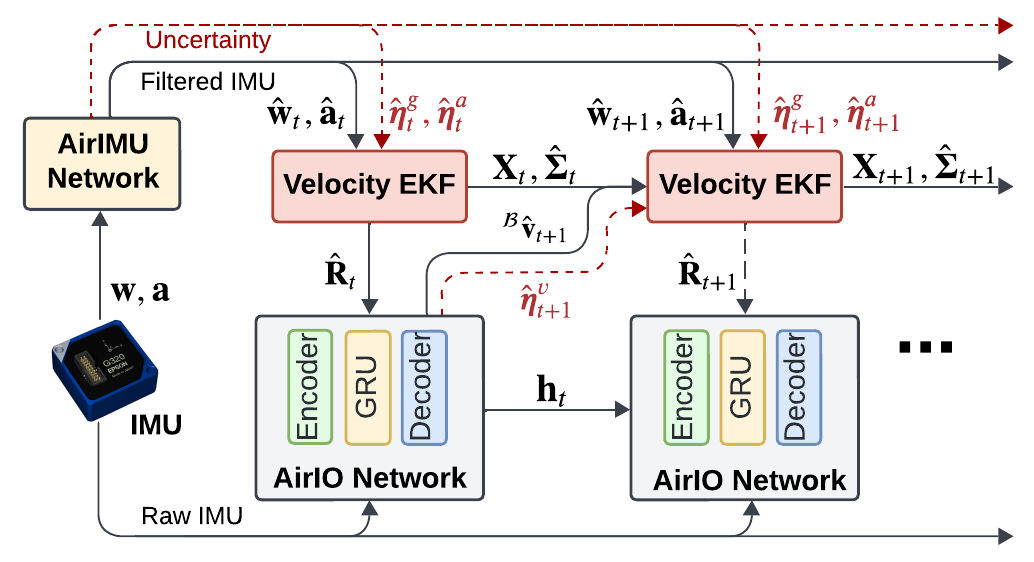}
    \caption{AirIO system pipeline. 
    A tightly coupled EKF that fuses the AirIO motion network with a learning-based IMU preintegration network AirIMU. Instead of using constant uncertainty parameters, we employ learned uncertainty $\hat{\boldsymbol{\eta}}^v$ from the AirIO network and the learned gyroscope uncertainty $\hat{\boldsymbol{\eta}}^g$ and accelerometer $\hat{\boldsymbol{\eta}}^a$ \bluetext{from the AirIMU} in our EKF system.}
    \label{fig:network-struct}
\vspace{-10pt}
\end{figure}

\bluetext{Different from the existing IO methods~\cite{liu2020tlio, CioffiRal2023} which employ a motion network to capture relative translation and maintain historic states to optimize relative pose for state pairs.}
\bluetext{AirIO network models the body-frame velocity. As a result, we only need to constrain the velocity of the current frame. which significantly reduces the number of state parameters and alleviates the need for frame culling.}
This setup isolates the current state of the drone from the previous state, which simplifies the EKF structure.
The full state of the filter is $\mathbf{X} = (\mathbf{R}_i, {}^\mathcal{G}\mathbf{v}_i, {}^\mathcal{G}\mathbf{p}_i, \mathbf{b}_{a_i}, \mathbf{b}_{g_i} )$, where $\mathbf{b}_{a_i}$ and $\mathbf{b}_{g_i}$ are the bias of the accelerometer and the gyroscope respectively. Following the common setup of the error-based filtering method, we define the error-state of the current filter as $\delta\mathbf{X} = (\delta{{}\boldsymbol{\xi}}_i, \delta{}^\mathcal{G}\mathbf{v}_i, \delta{}^\mathcal{G}\mathbf{p}_i, \delta\mathbf{b}_{a_i}, \delta\mathbf{b}_{g_i} )$. For the rotation error, we define $\delta{\boldsymbol{\xi}}_i = {\log}_{SO3}(\mathbf{R}_i \hat{\mathbf{R}}_i^{-1}) \in {\mathfrak{so}(3)}$, where ${\log}_{SO3}(\cdot)$ denotes logarithm map operator.

\textbf{AirIMU feature correction and uncertainty estimation}
To filter the noise and, more importantly, quantify the uncertainty of the IMU pre-integration, we leverage AirIMU~\cite{qiu2023airimu} to pre-process the raw IMU data. 
This model estimates the corrections of the gyroscope and the accelerometer, and also the uncertainty of the IMU sensor:
\begin{equation}
    \left(\hat{\boldsymbol{\sigma}}_{g_i}, \hat{\boldsymbol{\sigma}}_{a_i}\right) = g_{\boldsymbol{\theta}}\left(\mathbf{w}_i, \mathbf{a}_i\right), \ \ \  \left(\hat{\boldsymbol{\eta}}_{g_i}, \hat{\boldsymbol{\eta}}_{a_i}\right) = \Sigma_{\boldsymbol{\theta}}\left(\mathbf{w}_i, \mathbf{a}_i\right),
\end{equation}
where the $\mathbf{\theta}$ is the parameter of the neural network. The AirIMU network is trained by a differentiable IMU integrator and covariance propagator.
We later filtered the IMU measurement by the predicted correction $\hat{\mathbf{a}} = \mathbf{a} + \hat{\boldsymbol{\sigma}}_{a_i}, \hat{\mathbf{w}} = \mathbf{w} +  \hat{\boldsymbol{\sigma}}_{g_i}$. In the EKF system, the learned uncertainty $(\hat{\boldsymbol{\eta}}_{g_i}, \hat{\boldsymbol{\eta}}_{a_i})$ is leveraged in the sensor covariance modeling.

\textbf{Filter Propagation}
The kinematic motion model is:
\begin{equation}
    \begin{split}
    \label{formula: 2.integrate}
        &\mathbf{R}_{i+1} = \mathbf{R}_i \cdot \text{Exp}(\hat{\mathbf{w}}_i - \mathbf{b}_{g_i})  \Delta t, \\
        &\mathbf{v}_{i+1} = \mathbf{v}_i + \mathbf{R}_i (\hat{\mathbf{a}}_i - \mathbf{b}_{a_i}) {\Delta}t, \\
        &\mathbf{p}_{i+1} = \mathbf{p}_i + \mathbf{v}_i \Delta t + \frac{1}{2} \Delta t^2 \mathbf{R}_i( \hat{\mathbf{a}}_i - \mathbf{b}_{a_i}), \\
        &\mathbf{b}_{a_{i+1}} = \mathbf{b}_{a_{i}}, \mathbf{b}_{g_{i+1}} = \mathbf{b}_{g_{i}},
    \end{split}
\end{equation}
Here, the $\text{Exp}(\cdot)$ denotes mapping from the log space to exponential space.
The linearized propagation model is:
\begin{equation}
\delta \mathbf{X}_{i+1} = \mathbf{A}_i \cdot \delta \mathbf{X}_i + \mathbf{B}_i \cdot \mathbf{n}_i,
\end{equation}
where $\mathbf{n}_i = [ \hat{\boldsymbol{\eta}}_{g_i}, \hat{\boldsymbol{\eta}}_{a_i}, \boldsymbol{\eta}_{b_{g}}, \boldsymbol{\eta}_{b_{a}}]^T$. The $\hat{\boldsymbol{\eta}}_{g_i}, \hat{\boldsymbol{\eta}}_{a_i}$ are the learned IMU uncertainty from the AirIMU network \cite{qiu2023airimu}
with the $i$-th frame. 
The $\boldsymbol{\eta}_{b_{g}}, \boldsymbol{\eta}_{b_{a}}$ are the random walk uncertainty set as a constant across different frames. The corresponding covariance propagation of the state covariance $\mathbf{P}_{i+1}$ follows:
\begin{equation}
\mathbf{P}_{i+1} = \mathbf{A}_i \mathbf{P}_{i} \mathbf{A}_i + \mathbf{B}_i \mathbf{W}_i \mathbf{B}_i
\end{equation}
where the $\mathbf{W}_i$ is the covariance matrix of the $i$-th input, including the learned uncertainty of the IMU $ \hat{\boldsymbol{\eta}}_{g_i}, \hat{\boldsymbol{\eta}}_{a_i}$  and sensor bias random walk $\boldsymbol{\eta}_{b_{g}}, \boldsymbol{\eta}_{b_{a}}$.

\textbf{Filter Update} 
The measurement update is as follows:
\begin{equation}
    h(\mathbf{X}) = \hat{\mathbf{R}}_i^T \cdot \mathbf{v}_i =  \hat{\mathbf{v}}_i + \hat{ \boldsymbol{\eta}}_{v_i}.
\end{equation}
where $\boldsymbol{\eta}_{v_i}$ is a random variable that follow the Gaussian distribution $\mathcal{N}(0, \hat{\boldsymbol{\Sigma}}^v_i )$ learned by the network. The Jacobian matrix $\mathbf{H}$ is computed as:
\begin{equation}
\begin{split}
\mathbf{H}_{\delta v_i} &= \frac{\partial h(\mathbf{X})}{\partial \delta \mathbf{v}_i} =  \hat{\mathbf{R}}_i^T\\
\mathbf{H}_{\delta \xi_i} &= \frac{\partial h(\mathbf{X})}{\partial \delta \boldsymbol{\xi}_i} = \hat{\mathbf{R}}_i^T [\mathbf{v}_i]_\times
\end{split}
\end{equation}

Finally, the Kalman Gain and the update can be computed:
\begin{equation}
    \begin{split}
    &\mathbf{K} = \mathbf{P} \mathbf{H}^T(\mathbf{\bluetext{HPH}}^{\bluetext{T}} \bluetext{+} \bluetext{\hat{\boldsymbol{\Sigma}}}^{\bluetext{v}}_{\bluetext{i}} )^{\bluetext{-1}} \\
    &\mathbf{X} \leftarrow \mathbf{X} \oplus (\mathbf{K}(h(\mathbf{X}) -  \hat{\mathbf{v}}_i)) \\
    &\mathbf{P} \leftarrow (\mathbf{I} - \mathbf{KH})\mathbf{P}(\mathbf{I}-\mathbf{KH})^T + \mathbf{K}  \hat{\boldsymbol{ \Sigma}}^v_i \mathbf{K}^T
    \end{split}
\end{equation}
Operator $\oplus$ denotes the additional operation except for the orientation, where the update performs $\mathbf{R} \leftarrow \text{Exp} (\boldsymbol{\xi}) \cdot \mathbf{R}$.
\section{Experiment}
\begin{table*}[!t]
    \caption{The ATE (Unit: \meter) and RTE (Unit: \meter) on the Blackbird dataset. \textbf{Seen} are sequences where the training and testing datasets are derived from the same trajectory.  \textbf{Unseen} are those where the testing dataset includes sequences never used in training, demonstrating the model's generalizability. }
    \label{blackbird}
    \centering
    \resizebox{0.8\linewidth}{!}{
    \begin{tabular}{C{1cm}C{1.8cm}|C{.8cm}C{.8cm}C{.8cm}C{.8cm}C{.8cm}C{.8cm}C{.8cm}C{.8cm}C{.8cm}C{.8cm}C{.8cm}C{.8cm}C{.8cm}C{.8cm}}
        \toprule
        % \multirow{18}{*}{\rotatebox{90}{\textbf{Dataset Name}}} &
        \multirow{2}{*}{\textbf{}} &\multirow{2}{*}{\textbf{Seq.}} & \multicolumn{2}{c}{\textbf{Baseline$^\ddagger$}} & \multicolumn{2}{c}{\textbf{AirIMU$^\ddagger$}} & \multicolumn{2}{c}{\textbf{RoNIN$^\dagger$}}  & \multicolumn{2}{c}{\textbf{TLIO$^\dagger$}} & \multicolumn{2}{c}{\textbf{IMO$^\dagger$$^{\bluetext{\ast}}$ }} & \multicolumn{2}{c}{\textbf{AirIO Net}} & \multicolumn{2}{c}{\textbf{AirIO EKF}}\\
        &&  ATE  &  RTE &  ATE  &  RTE &  ATE  &  RTE &  ATE  &  RTE &  ATE  &  RTE &  ATE  &  RTE &  ATE  &  RTE \\
        \midrule
          % \multirow{6}{*}{\rotatebox{90}{ \textbf{Seen}}}

        % \raggedright &\textbf{Seen}  & \multicolumn{2}{c}{} & \multicolumn{2}{c}{} & \multicolumn{2}{c}{} & \multicolumn{2}{c}{} & \multicolumn{2}{c}{} & \multicolumn{2}{c}{} & \multicolumn{2}{c}{}\\
        
        % \raggedright\hspace{0.3cm}& 
        \multirow{6}{*}{{ \textbf{Seen}}}
        &Clover &\cellcolor{g9!100} 15.769 & \cellcolor{g9!100} 2.027 & \cellcolor{g7!100}{\bluetext{6.163}} & \cellcolor{g1!100}{\bluebf{0.294}}& \cellcolor{g6!100} 3.074 & \cellcolor{g7!100} 1.367 & \cellcolor{g4!100} 1.464 & \cellcolor{g6!100} 0.797 & \cellcolor{g2!100} 0.381 & \cellcolor{g4!100} 0.681 & \cellcolor{g3!100}{\bluetext{0.434}} & \cellcolor{g2!100} {\bluetext{0.368}} & \cellcolor{g1!100} {\bluebf{0.367}} & \cellcolor{g3!100} {\bluetext{0.391}} \\

        \raggedright\hspace{0.3cm} &Egg & \cellcolor{g9!100} 66.17 & \cellcolor{g9!100} 7.677 & \cellcolor{g7!100} \bluetext{13.293} & \cellcolor{g7!100} \bluetext{3.801} & \cellcolor{g6!100} 2.449 & \cellcolor{g6!100} 2.552 & \cellcolor{g4!100} 2.227 & \cellcolor{g4!100} 2.398 & \cellcolor{g3!100} 1.153 & \cellcolor{g3!100} 0.828 & \cellcolor{g2!100} \bluetext{0.713} & \cellcolor{g2!100}\bluetext{0.391} & \cellcolor{g1!100} \bluebf{0.408} & \cellcolor{g1!100} \bluebf{0.344} \\

        \raggedright\hspace{0.3cm}& halfMoon  & \cellcolor{g9!100} 19.165 & \cellcolor{g9!100} 3.386 & \cellcolor{g7!100} \bluetext{4.174} & \cellcolor{g6!100} \bluetext{0.746} & \cellcolor{g6!100} 1.263 & \cellcolor{g7!100} 0.864 & \cellcolor{g4!100} 0.956 & \cellcolor{g4!100} 0.475 & \cellcolor{g3!100} 0.761 & \cellcolor{g1!100} \textbf{0.24} & \cellcolor{g2!100} \bluetext{0.491} & \cellcolor{g3!100} \bluetext{0.256} & \cellcolor{g1!100} \bluebf{0.457} & \cellcolor{g2!100} \bluetext{0.253}\\
        
        \raggedright\hspace{0.3cm}& Star  &\cellcolor{g9!100} 20.49 & \cellcolor{g9!100} 4.556 & \cellcolor{g7!100} \bluetext{4.347} & \cellcolor{g4!100} \bluetext{1.933} & \cellcolor{g6!100} 3.15 & \cellcolor{g7!100} 3.266 & \cellcolor{g3!100} 0.68 & \cellcolor{g3!100} 0.784 & \cellcolor{g4!100} 2.13 & \cellcolor{g6!100} 3.066 & \cellcolor{g1!100} \bluebf{0.442} & \cellcolor{g2!100} \bluetext{0.401}& \cellcolor{g2!100} \bluetext{0.477} & \cellcolor{g1!100} \bluebf{0.341} \\
        
        \raggedright\hspace{0.3cm}& Winter &\cellcolor{g9!100} 15.781 & \cellcolor{g9!100} 2.395 & \cellcolor{g7!100} \bluetext{7.445} & \cellcolor{g4!100} \bluetext{0.743} & \cellcolor{g6!100} 1.031 & \cellcolor{g7!100} 1.089 & \cellcolor{g4!100} 0.616 & \cellcolor{g6!100} 0.755 & \cellcolor{g1!100} \textbf{0.219} & \cellcolor{g3!100} 0.206 & \cellcolor{g3!100} \bluetext{0.348} & \cellcolor{g1!100} \bluebf{0.147} & \cellcolor{g2!100} \bluetext{0.307} & \cellcolor{g2!100} \bluetext{0.164} \\
        % \midrule
        \cmidrule{2-16}
        &\textbf{Avg.}& \cellcolor{g9!100} 27.475 & \cellcolor{g9!100} 4.008 & \cellcolor{g7!100} \bluetext{7.084} & \cellcolor{g6!100} \bluetext{1.503} & \cellcolor{g6!100} 2.193 & \cellcolor{g7!100} 1.828 & \cellcolor{g4!100} 1.189 & \cellcolor{g4!100} 1.042 & \cellcolor{g3!100} 0.929 & \cellcolor{g3!100} 1.004 & \cellcolor{g2!100} \bluetext{0.486} & \cellcolor{g2!100} \bluetext{0.312} & \cellcolor{g1!100} \bluebf{0.403} & \cellcolor{g1!100} \bluebf{0.299} \\
        
        \midrule

        % \raggedright &\textbf{Unseen} & \multicolumn{2}{c}{} & \multicolumn{2}{c}{} & \multicolumn{2}{c}{} & \multicolumn{2}{c}{} & \multicolumn{2}{c}{} & \multicolumn{2}{c}{} & \multicolumn{2}{c}{} \\
        
        % \raggedright\hspace{0.3cm} 
        \multirow{6}{*}{{ \textbf{Unseen}}}
        &Ampersand &\cellcolor{g9!100} 41.138 & \cellcolor{g6!100} 5.026 & \cellcolor{g7!100}{\bluetext{24.738}} & \cellcolor{g4!100} {\bluetext{4.379}} & \cellcolor{g4!100} 5.467 & \cellcolor{g7!100} 5.321 & \cellcolor{g3!100} 4.665 & \cellcolor{g3!100} 4.078 & \cellcolor{g6!100} 17.664 & \cellcolor{g9!100} 10.039 & \cellcolor{g1!100} {\bluebf{2.303}} & \cellcolor{g1!100} {\bluebf{1.145}} & \cellcolor{g2!100} \bluetext{2.334} & \cellcolor{g2!100} \bluetext{1.154} \\
        
        \raggedright\hspace{0.3cm}& Sid &\cellcolor{g6!100} 13.108 & \cellcolor{g3!100} 2.055 & \cellcolor{g7!100} \bluetext{16.832} & \cellcolor{g4!100} \bluetext{2.131} & \cellcolor{g9!100} 18.581 & \cellcolor{g9!100} 10.665 & \cellcolor{g3!100} 7.323 & \cellcolor{g7!100} 8.427 & \cellcolor{g4!100} 9.967 & \cellcolor{g6!100} 6.992 & \cellcolor{g1!100} \bluebf{0.717} & \cellcolor{g2!100} \bluetext{0.527} & \cellcolor{g2!100} \bluetext{0.874} & \cellcolor{g1!100} \bluebf{0.49} \\
        
        \raggedright\hspace{0.3cm}& Oval & \cellcolor{g7!100} 15.365 & \cellcolor{g6!100} 2.4 & \cellcolor{g6!100} \bluetext{8.312} & \cellcolor{g4!100} \bluetext{1.553} & \cellcolor{g9!100} 20.399 & \cellcolor{g9!100} 12.12 & \cellcolor{g3!100} 1.276 & \cellcolor{g3!100} 1.045 & \cellcolor{g4!100} 5.67 & \cellcolor{g7!100} 2.863 & \cellcolor{g2!100} \bluetext{0.976} & \cellcolor{g2!100} \bluetext{0.583} & \cellcolor{g1!100} \bluebf{0.828} & \cellcolor{g1!100} \bluebf{0.54} \\
        
        \raggedright\hspace{0.3cm}& Sphinx &\cellcolor{g6!100} 3.429 & \cellcolor{g6!100} 2.713 & \cellcolor{g4!100} \bluetext{2.828} & \cellcolor{g3!100} \bluetext{2.024} & \cellcolor{g9!100} 11.537 & \cellcolor{g9!100} 7.762 & \cellcolor{g3!100} 2.005 & \cellcolor{g4!100} 2.408 & \cellcolor{g7!100} 5.629 & \cellcolor{g7!100} 5.395 & \cellcolor{g1!100} \bluebf{1.145} & \cellcolor{g1!100} \bluebf{1.008} & \cellcolor{g2!100} \bluetext{1.178} & \cellcolor{g2!100} \bluetext{1.033} \\
        
        \raggedright\hspace{0.3cm}& BentDice &\cellcolor{g7!100} 28.307 & \cellcolor{g4!100} 2.342 & \cellcolor{g9!100} \bluetext{54.261} & \cellcolor{g6!100} \bluetext{4.39} & \cellcolor{g6!100} 11.453 & \cellcolor{g9!100} 6.792 & \cellcolor{g3!100} 2.119 & \cellcolor{g3!100} 1.747 & \cellcolor{g4!100} 6.143 & \cellcolor{g7!100} 4.519 & \cellcolor{g2!100} \bluetext{1.360} & \cellcolor{g2!100} \bluetext{1.000} & \cellcolor{g1!100} \bluebf{1.331} & \cellcolor{g1!100} \bluebf{0.955} \\
        
        % \midrule
        \cmidrule{2-16}
        &\textbf{Avg.} &\cellcolor{g7!100} 20.269 & \cellcolor{g4!100} 2.907 & \cellcolor{g9!100} \bluetext{21.403} & \cellcolor{g3!100} \bluetext{2.896} & \cellcolor{g6!100} 13.487 & \cellcolor{g9!100} 8.532 & \cellcolor{g3!100} 3.478 & \cellcolor{g6!100} 3.541 & \cellcolor{g4!100} 9.015 & \cellcolor{g7!100} 5.962 & \cellcolor{g1!100} {\bluebf{1.300}} & \cellcolor{g2!100} {\bluetext{0.853}}& \cellcolor{g2!100} {\bluetext{1.309}} & \cellcolor{g1!100} {\bluebf{0.834}} \\
        \bottomrule
        \multicolumn{9}{l}{\bluetext{$^{\ast}$ Leverage thrust information as the input data for training and inference.}}\\
        % \multicolumn{12}{l}{$^\dagger$ Leverage ground truth orientation to transform the input data.}\\
        % \multicolumn{9}{l}{$^\ddagger$ Dead-reckoning IMU pre-integration.}
        \multicolumn{12}{l}{$^\dagger$ Leverage ground truth orientation to transform the input data. $^\ddagger$ \bluetext{Dead-reckoning IMU pre-integration.}}
    \end{tabular} 
    }
    \vspace{-1em}
\end{table*}

\subsection{Experiment Setup}
% EuRoC micro aerial vehicle datasets are a widely used benchmark for evaluating for Odometry. 
% The Blackbird dataset is an aggressive indoor flight dataset collected using a quadrotor platform. 
\noindent\textbf{Datasets} We evaluate the proposed approach on two quadrotor datasets: the real-world EuRoC dataset~\cite{burri2016euroc} and the challenging drone racing dataset Blackbird~\cite{antonini2018blackbird}. EuRoC datasets are a widely used benchmark for evaluating odometry.
\bluetext{Blackbird dataset is an indoor flight dataset, presenting more aggressive maneuvers and high-speed dynamics.}
For ablation studies, we additionally include our custom simulated dataset collected using the NVIDIA IsaacSim simulator with Pegasus autopilot \cite{Jacinto2024pegasus}. 
The Pegasus dataset also presents diverse maneuvers but in the ideal environment without external disturbance, allowing us to work with cleaner data for ground-truth accuracy. 

\noindent\textbf{Baseline} Several state-of-the-art methods are selected for comparison. For model-based IO, we include \textit{Baseline}~\cite{forster2015imu}, the IMU preintegration approach in raw IMU data, and \textit{AirIMU}~\cite{qiu2023airimu}, a hybrid method that corrects the raw IMU noise with a data-driven method and integrates it with the IMU kinematic function. 
For learning-based IO, we compare against an end-to-end trained model \textit{RoNIN}~\cite{herath2020ronin} and an EKF-fused algorithm \textit{TLIO}~\cite{liu2020tlio}. 
For learning-based IO methods designed for UAVs, we evaluate \textit{IMO}~\cite{CioffiRal2023}, a control signal-embedded IO specifically developed for drone racing, which incorporates additional thrust signals to better capture drone motion.

\noindent\textbf{Evaluation Metrics} We use the following metrics for evaluation: 1) \textbf{Absolute Translation Error} (ATE, $\si{\meter}$): the average Root Mean Squared Error (RMSE) between the estimated and ground-truth positions over all time points: 
\begin{equation}
    \text{ATE} =\sqrt{\frac{1}{N}\sum^N_{i=1}{\left\| \mathbf{p}_{i} - \hat{\mathbf{p}}_{i}\right\|_2^2}},
\end{equation}
2) \textbf{Relative Translation Error} (RTE, $\si{\meter}$): the average RMSE of the relative displacements over predefined time intervals.
\begin{equation}
    \text{RTE} = \sqrt{ \frac{1}{N}\sum^N_{i=1}\left\| \mathbf{p}_{i+\Delta t} - \mathbf{p}_{i} - \mathbf{R}_{i}\hat{\mathbf{R}}_{i}^T \left(\hat{\mathbf{p}}_{i+\Delta t} - \hat{\mathbf{p}}_{i}\right)\right\|^2_2},
\end{equation}
Our evaluation adopts a 5-second interval configuration.

We define the improvement percentage as the relative reduction in error compared to the baseline. Specifically, it is computed as the difference between the error of baseline and the proposed method, divided by the baseline error.

\noindent\textbf{Training Details} We use the Adam optimizer with an initial learning rate of 0.001. A learning rate scheduler is implemented following the ReduceLROnPlateau, with a patience of 5 epochs and a decay factor of 0.2. The batch size is 128. Our training and testing window size is 1000 frames (equal to 5 seconds in the EuRoC dataset). In the CNN encoder, we include a dropout layer with $p=0.5$ to reduce overfitting.

\subsection{Results}
\subsubsection{Evaluation on Blackbird Dataset} 
We employ the same sequences from Blackbird as used in IMO \cite{CioffiRal2023} and adopt their same train-test split configuration. We name these sequences as \textbf{SEEN} sequences because the training and testing data are from the same sequence. To evaluate generalization, we select five additional trajectories from BlackBird dataset as \textbf{UNSEEN} sequences, which are excluded from the training set. More details on separation are in \aref{Appendix:Seen-Unseen}.

\begin{figure}[h]
    \centering
    \includegraphics[width=1\linewidth]{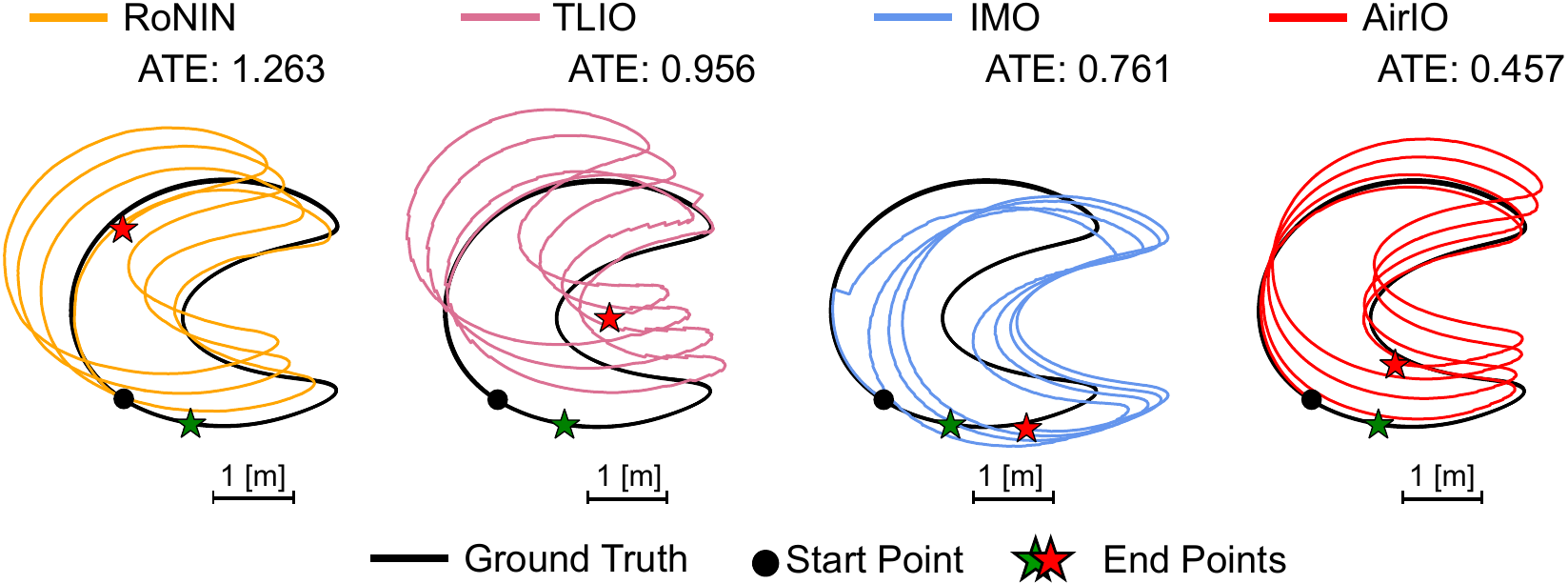}
    {\caption{Trajectories of \textbf{SEEN} sequence \texttt{halfMoon} from the Blackbird dataset. AirIO, relying solely on IMU, outperforms IMO by 30\% in ATE. }}
    \label{fig:seen_halfmoon}
\end{figure}
For the seen sequences, as shown in \tref{blackbird}, RoNIN and TLIO fail to capture the aggressive drone maneuvers as they are designed for pedestrian navigation with clear slow motion patterns. 
% Unlike pedestrian datasets that demonstrate clear motion patterns, quadrotor motion, especially from high-speed quadrotors, is more unpredictable. 
AirIO significantly outperforms these baselines, improving by \bluetext{66.1\% in ATE and 71.3\% in RTE} over TLIO. 
\begin{figure}[h]
    \centering
    \includegraphics[width=1\linewidth]{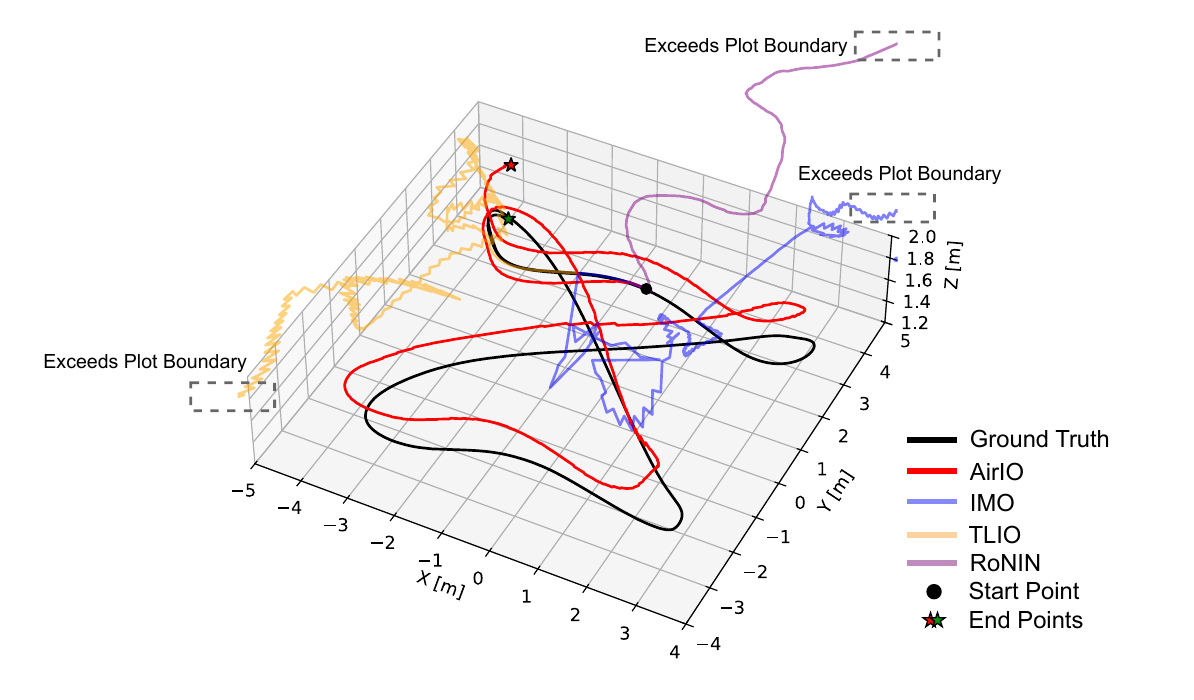}
    \caption{Trajectories of \textbf{UNSEEN} sequence \texttt{sid} from the Blackbird dataset by RoNIN, TLIO, IMO, and AirIO.}
    % The red trajectory, estimated by AirIO, shows superior performance with an ATE of 0.686 $\meter$.
    
    % In contrast, other methods demonstrate immediate drift, with ATE values ranging from 7.323 to 18.681 $\meter$.
    
    \label{fig:unseen_sid}
\end{figure}
Furthermore, AirIO exceeds IMO by \bluetext{56.6\% in ATE and 70.2\% in RTE }relying solely on IMU measurement without integrating external thrust information. \bluetext{We also show that our algorithm can outperform the multi-sensor fusion algorithms like the visual-inertial odometry on the drone racing data. For more details, please check \aref{Appendix:VIO}}.
% AirIMU relies solely on IMU data without incorporating thrust or other additional inputs. 

% IMO improves by 21.86\% in ATE and 3.61\% in RTE over TLIO. The previous work~\cite{kaufmann2022benchmark} has demonstrated that collective thrust is a preferred control input modality for agile quadrotor flight, and collective thrust is more robust to changes in dynamics.

% Surprisingly, we found that all existing methods, especially IMO, excel on {SEEN} sequences but experience serious drift on {UNSEEN} sequences as shown in \ref{fig:unseen_sid}. 
In addition, it is worth noting that AirIO demonstrates remarkable generalizability to unseen sequences, as shown in \fref{fig:unseen_sid}. While other methods suffer from significant drift, AirIO consistently maintains precise estimations.
The baseline methods tend to overfit the training data, performing well within the training distribution but failing to generalize to UNSEEN data. Compared to IMO, AirIO shows a substantial improvement of \bluetext{85.5\% in ATE and 86.0\% in RTE}. 

% AirIO  while other existing methods experience serious drift. To be more specific, when compared to TLIO, AirIO shows an improvement of 65.3\% in ATE and 74.3\% in RTE. Against IMO, the improvements are even more observable with 86.6\% in ATE and 84.7\% in RTE. Among them, the integration of EKF results in a 5.9\% enhancement in ATE. The results illustrate the generalizability of AirIO. On the contrary, existing methods only excel on seen sequences,  Overall, the utilization of body coordinate frame and attitude information assists the model in handling the faster and more complex dynamics of the Blackbird dataset.

\subsubsection{Evaluation on EuRoC Dataset}
% The EuRoC dataset contends 11 trajectories, including dynamic flights. 
% We use 6 trajectories for training, specifically $MH\_01$, $MH\_03$, $MH\_05$, $V1\_02$, $V2\_01$, $V2\_03$. For the testing set, we select the remaining trajectories: $MH\_02$, $MH\_04$,  $V1\_01$, $V2\_02$, $V1\_03$. In this way we cover a range of flight dynamics and environments, proving a comprehensive evaluation of our model's performance.
% In \tref{euroc}, we show the ATE and RTE estimated by RoNIN, TLIO, and AirIO. 
We then evaluate the proposed method on EuRoC dataset, as shown in \tref{euroc}. AirIO outperforms both RoNIN and TLIO by 52.9\% and 54.4\% in ATE, respectively. Thanks to effective uncertainty modeling, AirIO leverages uncertainties learned from both the IMU kinematic model and motion network, ensuring optimal data fusion and improved performance.
% AirIO network shows a 43.0\% and 44.8\% improvement in ATE and a 52.4\% and 60.8\% improvement in RTE compared to RoNIN and TLIO, respectively. 
After integrating EKF, AirIO further improves accuracy, showing an average additional 17.4\% improvement in ATE. 

% The success of AirIO can be attributed to its learning of the drone's body coordinate frame and its attitude information encoding. 
% They aid the AirIO model in better understanding drone dynamics and enhance the AirIO's ability to learn the complex dynamics of drones.
\begin{table}[h]
    \caption{The ATE (Unit: \meter) and RTE (Unit: \meter) on EuRoC dataset.}
    \label{euroc}
    \centering
\resizebox{1\linewidth}{!}{
    \begin{tabular}{C{2cm}|C{.8cm}C{.8cm}C{.8cm}C{.8cm}C{.8cm}C{.8cm}C{.8cm}C{.8cm}}
        \toprule
        \multirow{2}{*}{\textbf{Seq.}} & \multicolumn{2}{c}{\textbf{RoNIN}$^\dagger$} & \multicolumn{2}{c}{\textbf{TLIO}$^\dagger$} & \multicolumn{2}{c}{\textbf{AirIO Net}} & \multicolumn{2}{c}{\textbf{AirIO EKF}}\\
        &  ATE  &  RTE &  ATE  &  RTE &  ATE  &  RTE &  ATE  &  RTE\\
        
        \midrule
         MH02 &\cellcolor{g5!100} 5.902 & \cellcolor{g5!100} 2.121 & \cellcolor{g7!100} 7.281 & \cellcolor{g7!100} 2.451 & \cellcolor{g3!100} 4.917 & \cellcolor{g1!100} \textbf{0.936} & \cellcolor{g1!100} \textbf{2.478} & \cellcolor{g3!100} 0.987 \\
         
         MH04 &\cellcolor{g5!100} 8.586 & \cellcolor{g5!100} 4.542 & \cellcolor{g7!100} 8.626 & \cellcolor{g7!100} 5.498 & \cellcolor{g3!100} 2.726 & \cellcolor{g3!100} 1.093 & \cellcolor{g1!100} \textbf{2.308} & \cellcolor{g1!100} \textbf{1.005} \\

         V103 & \cellcolor{g3!100} 3.240 & \cellcolor{g5!100} 1.695 & \cellcolor{g7!100} 7.863 & \cellcolor{g7!100} 2.580 & \cellcolor{g5!100} 3.844 & \cellcolor{g1!100} \textbf{1.519} & \cellcolor{g1!100} \textbf{3.05} & \cellcolor{g3!100} 1.552 \\

         V202 &\cellcolor{g7!100} 7.445 & \cellcolor{g5!100} 2.303 & \cellcolor{g5!100} 6.260 & \cellcolor{g7!100} 2.783 & \cellcolor{g3!100} 4.823 & \cellcolor{g1!100} \textbf{1.303} & \cellcolor{g1!100} \textbf{4.206} & \cellcolor{g3!100} 1.313 \\
         
         V101 &\cellcolor{g7!100} 8.576 & \cellcolor{g5!100} 1.918 & \cellcolor{g5!100} 4.814 & \cellcolor{g7!100} 1.946 & \cellcolor{g1!100} \textbf{2.917} & \cellcolor{g3!100} 1.137 & \cellcolor{g3!100} 3.844 & \cellcolor{g1!100} \textbf{1.103} \\
        \midrule
        \textbf{Avg.} &\cellcolor{g5!100} 6.75 & \cellcolor{g5!100} 2.516 & \cellcolor{g7!100} 6.969 & \cellcolor{g7!100} 3.052 & \cellcolor{g3!100} 3.846 & \cellcolor{g3!100} 1.198 & \cellcolor{g1!100} \textbf{3.177} & \cellcolor{g1!100} \textbf{1.192} \\
    
        \bottomrule
        \multicolumn{9}{l}{$^\dagger$ Leverage ground truth orientation to transform the input data for inference.}\\

    \end{tabular} 
    }
    \vspace{-2em}
\end{table}

\subsubsection{\bluetext{Real time deployment and Analysis}}
\bluetext{We evaluate the real-time performance of our algorithm on both an NVIDIA Jetson Orin AGX and a desktop with a low-end RTX 2060 GPU, using an IMU operating at 200 Hz. On Orin, AirIO achieves an average runtime of 74.23 ms, while on the desktop, it runs in 28.92 ms. These results indicate the feasibility of our algorithm for real drone deployment.}

\section{Ablation Study}
\label{Sec: Ablation}
We perform an ablation study to investigate the impact of different feature representations. Experiments with the following input representation scenarios are conducted on the real-world dataset EuRoC and the simulated dataset Pegasus. 
\begin{enumerate}[leftmargin=*, label=\roman*.]
    \item \textbf{Body}: The network takes body-frame IMU measurements defined in (\ref{eq:body}) as input.\label{abla:body}
    \item \textbf{Global}: The network takes global-frame IMU measurements as input. The global-frame IMU is transformed from raw IMU data, as defined in (\ref{eq:global}). \label{abla:global}
    \item \textbf{Body} + \textbf{Attitude}: Building upon \iref{abla:body}, the drone orientation is encoded as auxiliary inputs to the network. \label{abla:bwattitude} 
    \item \textbf{Global} + \textbf{Attitude}: Building upon \iref{abla:global}, the drone orientation is encoded as auxiliary inputs to the network.\label{abla:gwattitude} 
    \item \textbf{Body} -- $\mathbf{R}^\top_i {}^\mathcal{G}\mathbf{g}$: 
    Building upon \iref{abla:body}, we consider the motion acceleration by removing the rotation-coupled gravity term $\mathbf{R}^\top_i {}^\mathcal{G}\mathbf{g}$ from (\ref{eq:body}).\label{abla:bw/og}
    % To verify the gravity term's role in enhancing rotation information and attitude observability, we implemented a body-frame version without it. 
    \item \textbf{Global} -- ${}^\mathcal{G}\mathbf{g}$: Building upon \iref{abla:global}, we consider the motion acceleration by removing the gravity term ${}^\mathcal{G}\mathbf{g}$ from (\ref{eq:global}).\label{abla:gw/og}
\end{enumerate}
% \begin{enumerate}
%     \item \textbf{Global v.s. Body}: 
%     The raw IMU data is transfrom from the quadrotor body frame to the global frame (Global), which is commonly used in existing IO algorithms.
%     \item \textbf{Impact of the gravity}:
%     \item \textbf{Attitude encoding}:
% \end{enumerate}
To ensure a fair comparison, scenarios \iref{abla:body}, \iref{abla:bwattitude}, and \iref{abla:bw/og} leverage ground-truth orientation to transform estimated velocity into global frame.

\tref{table:ablation} shows the results of the ablation study for all variants. More results can be found in \aref{Appendix:Pegasus_eval}. From these results, we address the following questions:
\begin{figure}[h]
    \centering
    \includegraphics[width=1\linewidth]{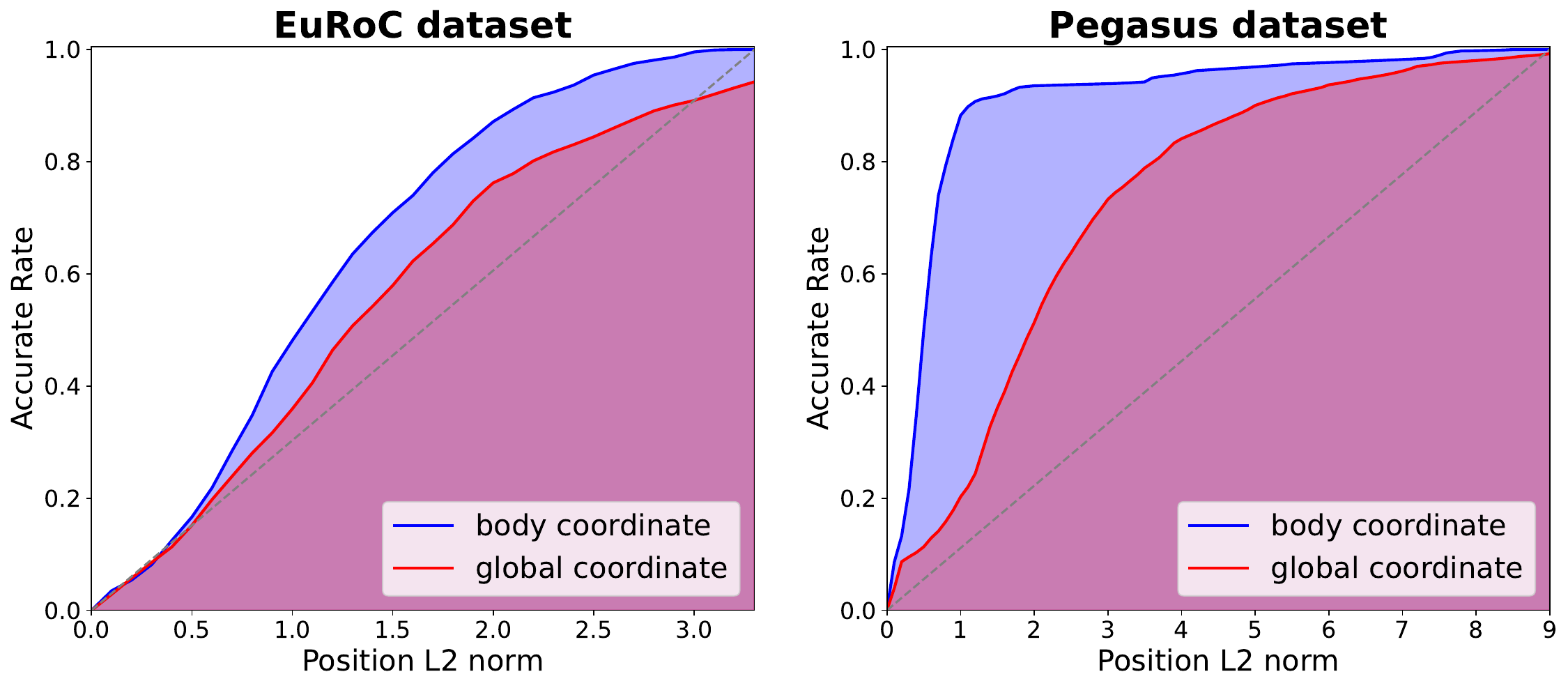}
    \caption{Accuracy AUC of EuRoC dataset \textbf{(Left)} and Pegasus dataset \textbf{(Right)}.}
    \label{fig:auc}
    \vspace{-5pt}
\end{figure}
\subsection{How Effective is the Body-frame Representation?} 
% \textbf{\iref{abla:HACF}{HACF}}, 
Compared to \iref{abla:global}{Global}, \iref{abla:body}{Body} demonstrates significantly higher precision on the Pegasus. This new input representation results in an average ATE improvement of 74.1\%. Scenario \iref{abla:body} on EuRoC dataset and Blackbird Unseen sequences exhibit similar trends.

We further evaluate the system performance under two representation using the Accuracy Area Under the Curve (AUC) metric, which quantifies the performance across various error thresholds. Specifically, we calculate the relative position error over a 5-second interval.
As shown in \fref{fig:auc}, on both datasets, scenario \iref{abla:body} achieves higher accuracy across all error thresholds and reaches near-perfect accuracy much more quickly. 
These results highlight the effectiveness of body-frame representation which efficiently captures implicit attitude information of IMU data. Simply encoding the body frame IMU features improves the system performance. 

%helps the network learn the drone's dynamics. %Moreover, we find that combining the body-frame representation with the attitude information \textbf{\iref{abla:bwattitude}{Body + Attitude}} is better than \textbf{\iref{abla:body}{Body}}. 

\begin{table}[!t]
    \normalsize
    \caption{Ablation study on the EuRoC and Pegasus datasets comparing different feature representations. Evaluation metric: ATE (Unit: $\meter$).}
    \label{table:ablation}
    \centering
    \resizebox{1\linewidth}{!}{
    \begin{tabular}{C{0.1cm}C{1.8cm}|C{1.25cm}C{1.25cm}C{1.25cm}C{1.25cm}C{1.25cm}C{1.25cm}C{1.25cm}}
\toprule
         \multirow{2}{*}{\textbf{}} &\multirow{2}{*}
        {\textbf{Seq.}}& \multirow{2}{*}{\shortstack{\textbf{Global} (\ref{eq:global}) \\ $- {}^\mathcal{G}\mathbf{g}$}}& \multirow{2}{*}{\textbf{Global}}& \multirow{2}{*}{\shortstack{\textbf{Body} (\ref{eq:body})\\ $- \mathbf{R}_i^T {}^\mathcal{G}\mathbf{g}$}}& \multirow{2}{*}{\shortstack{\textbf{Global} \\ \textbf{$+$Attitude}}} & \multirow{2}{*}{\textbf{Body}} & \multirow{2}{*}{\shortstack{\textbf{Body} \\ \textbf{$+$Attitude}}}\\ 
        && & & & & & \\

         \midrule

        % \raggedright \textbf{EuRoC}  & 
        % \multicolumn{1}{c}{} & \multicolumn{1}{c}{} & \multicolumn{1}{c}{} & \multicolumn{1}{c}{} & \multicolumn{1}{c}{} & \multicolumn{1}{c}{} \\
   
         \multirow{6}{*}{{ \rotatebox{90}{\textbf{EuRoC}}}}&MH02& \cellcolor{g7!100} 18.048 & \cellcolor{g6!100} 17.892 & \cellcolor{g4!100} 16.136 & \cellcolor{g3!100} 9.522 & \cellcolor{g2!100} 4.537 & \cellcolor{g1!100} \textbf{2.531} \\
         
         &MH04 & \cellcolor{g7!100} 15.904 & \cellcolor{g4!100} 6.895 & \cellcolor{g6!100} 8.977 & \cellcolor{g3!100} 4.572 & \cellcolor{g2!100} 3.561 & \cellcolor{g1!100} \textbf{2.250} \\
         
         &V103 & \cellcolor{g6!100} 7.007 & \cellcolor{g7!100} 7.086 & \cellcolor{g3!100} 6.586 & \cellcolor{g2!100} 4.068 & \cellcolor{g4!100} 6.824 & \cellcolor{g1!100} \textbf{3.107} \\
         
         &V202 & \cellcolor{g6!100} 9.662 & \cellcolor{g2!100} 3.496 & \cellcolor{g4!100} 6.161 & \cellcolor{g7!100} 10.274 & \cellcolor{g1!100} \textbf{3.248} & \cellcolor{g3!100} 4.310 \\
         &V101  & \cellcolor{g6!100} 8.142 & \cellcolor{g7!100} 15.11 & \cellcolor{g3!100} 7.082 & \cellcolor{g4!100} 7.117 & \cellcolor{g2!100}  5.478 & \cellcolor{g1!100} \textbf{4.287} \\
          \cmidrule(l){2-8}
         &\textbf{Avg.} &\cellcolor{g7!100} 11.753 & \cellcolor{g6!100} 10.096 & \cellcolor{g4!100} 8.988 & \cellcolor{g3!100} 7.110 & \cellcolor{g2!100}  4.730 & \cellcolor{g1!100} \textbf{3.297} \\

\midrule
\midrule
        % \raggedright \textbf{Pegasus}  & \multicolumn{1}{c}{} & \multicolumn{1}{c}{} & \multicolumn{1}{c}{} & \multicolumn{1}{c}{} & \multicolumn{1}{c}{}  \\   
        \multirow{4}{*}{{ \rotatebox{90}{\textbf{Pegasus}}}}& 
        TEST\_1&  \cellcolor{g7!100} 22.115 & \cellcolor{g6!100} 9.689 & \cellcolor{g4!100} 9.534 & \cellcolor{g3!100} 6.151 & \cellcolor{g2!100} 5.225 & \cellcolor{g1!100} \textbf{3.418} \\
        &TEST\_2 & \cellcolor{g7!100} 15.745 & \cellcolor{g6!100} 15.185 & \cellcolor{g4!100} 14.729 & \cellcolor{g2!100} 5.373 & \cellcolor{g1!100} \textbf{4.489} & \cellcolor{g3!100} 5.719 \\
        &TEST\_3 & \cellcolor{g6!100} 19.467 & \cellcolor{g7!100} 26.960 & \cellcolor{g2!100} 3.580 & \cellcolor{g4!100} 14.545 & \cellcolor{g3!100} 3.734 & \cellcolor{g1!100} \textbf{1.786} \\
         \cmidrule(l){2-8}
       &\textbf{Avg.} & \cellcolor{g7!100} 19.109 & \cellcolor{g6!100} 17.278 & \cellcolor{g4!100} 9.281 & \cellcolor{g3!100} 8.690 & \cellcolor{g2!100} 4.483 & \cellcolor{g1!100} \textbf{3.641} \\

\midrule
\midrule
        % \raggedright \textbf{Blackbird}  & \multicolumn{1}{c}{} & \multicolumn{1}{c}{} & \multicolumn{1}{c}{} & \multicolumn{1}{c}{} & \multicolumn{1}{c}{}  \\ 
         \multirow{6}{*}{{ \rotatebox{90}{\textbf{Blackbird}}}}&
        Ampersand &\cellcolor{g7!100} 21.93 & \cellcolor{g4!100} 17.863 & \cellcolor{g6!100} 18.639 & \cellcolor{g3!100} 14.487 & \cellcolor{g1!100} \textbf{1.977} & \cellcolor{g2!100} 2.492 \\
        &Sid &\cellcolor{g7!100} 11.866 & \cellcolor{g6!100} 7.915 & \cellcolor{g4!100} 5.686 & \cellcolor{g3!100} 2.307 & \cellcolor{g2!100} 1.108 & \cellcolor{g1!100} \textbf{0.651} \\
        &Oval&\cellcolor{g6!100} 7.212 & \cellcolor{g4!100} 6.743 & \cellcolor{g3!100} 1.656 & \cellcolor{g7!100} 7.354 & \cellcolor{g2!100} 1.353 & \cellcolor{g1!100} \textbf{0.913} \\
        &Sphinx&\cellcolor{g6!100} 3.327 & \cellcolor{g7!100} 3.572 & \cellcolor{g3!100} 2.117 & \cellcolor{g4!100} 2.784 & \cellcolor{g2!100} 1.266 & \cellcolor{g1!100} \textbf{1.163} \\
       & BentDice&\cellcolor{g7!100} 15.5 & \cellcolor{g4!100} 8.876 & \cellcolor{g3!100} 6.989 & \cellcolor{g6!100} 9.626 & \cellcolor{g2!100} 1.915 & \cellcolor{g1!100} \textbf{1.249} \\
       \cmidrule(l){2-8}
      &\textbf{Avg.}&\cellcolor{g7!100} 11.967 & \cellcolor{g6!100} 8.994 & \cellcolor{g3!100} 7.018 & \cellcolor{g4!100} 7.312 & \cellcolor{g2!100} 1.524 & \cellcolor{g1!100} \textbf{1.294} \\

        \bottomrule
    \end{tabular} 
    }
    \vspace{-1em}
\end{table}

\subsection{How Effective is the Attitude Encoding?}

\begin{figure}[h]
    \centering
    \includegraphics[width=1\linewidth]{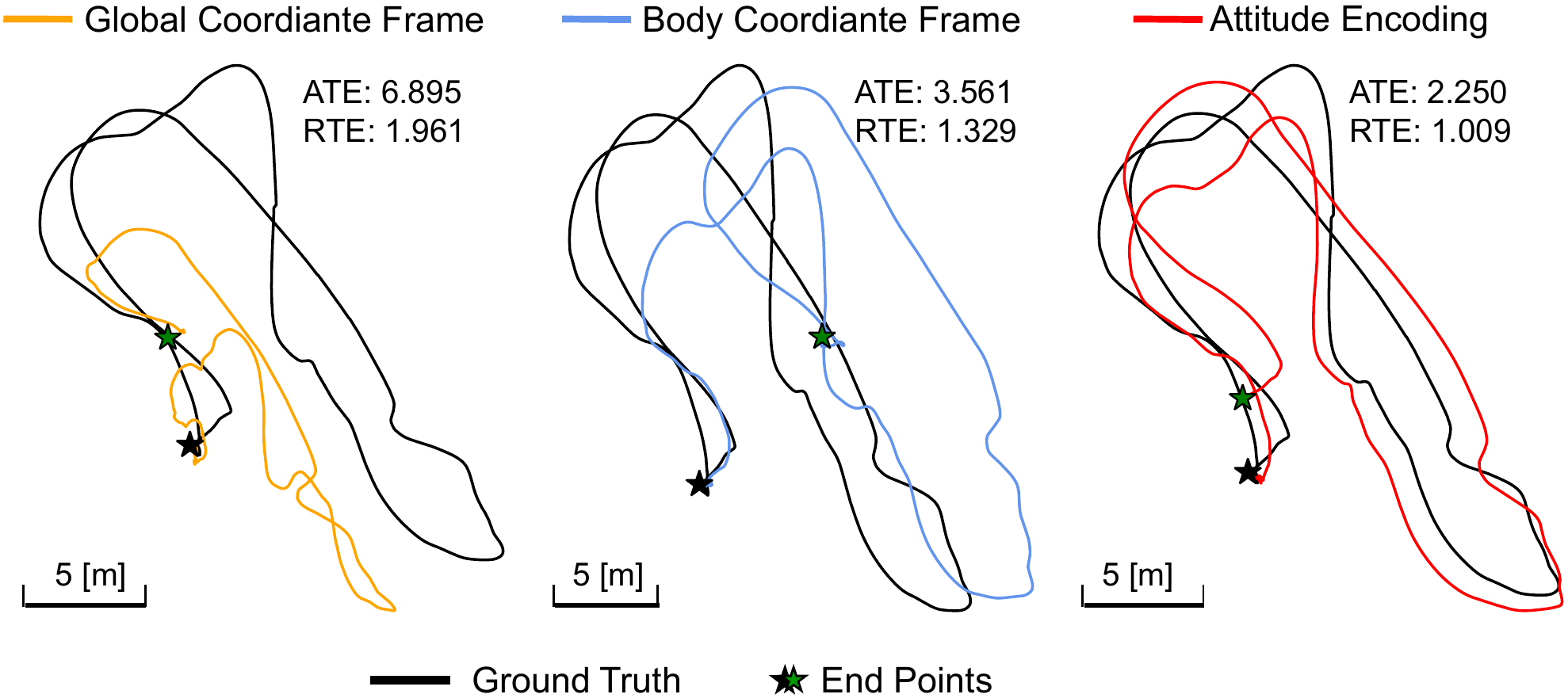}
    \caption{Performance of \texttt{MH\_04\_difficult} in EuRoC dataset across different representations. \textbf{\iref{abla:global}{Global}} \textbf{(Left)} is suboptimal. Using IMU data in \textbf{\iref{abla:body}{Body}} \textbf{(Center)} improves significantly. Encoding UAV orientation \textbf{\iref{abla:bwattitude}{Body + Attitude}} \textbf{(Right)} further improves the performance.}
    \label{fig:euroc_ablation}
    \vspace{-5pt}
\end{figure}

As illustrated in \fref{fig:euroc_ablation}, on the EuRoC dataset, \iref{abla:bwattitude}{Body + Attitude} achieves a further 30.3\% improvement over the scenario \iref{abla:body}, outperforming scenario \iref{abla:global} by 67.3\% in ATE.
Additionally, we evaluate attitude encoding on global-frame representation \iref{abla:gwattitude}{Global + Attitude}. This also yields an improvement of approximately 29.6\% over scenario \iref{abla:global} on EuRoC dataset, validating the efficiency of encoding attitude information.
Encoding the drone's attitude information enhances the network's ability for state estimation.
% In the Pegasus dataset, incorporating attitude information results in additional improvements of approximately 22.0\% in ATE. Compared to the global coordinate frame, \textbf{Attitude} leads to a significant improvement of 78.9\% in ATE. 
% Similarly, in the EuRoC dataset, attitude information improves ATE by 30.3\% compared to \textbf{Body} and by 67.3\% compared to the \textbf{Global}. 

\subsection{Shall We Keep Gravity in the Feature Representation?}
Many learning-based IO methods remove gravity from the IMU encoding \cite{bajwa2024dive}. However, as discussed in \sref{sec:coordinate}, gravitational acceleration is a critical component for body-frame representation as it explicitly embeds drone rotation through the term $\mathbf{R}^T_i {}^\mathcal{G}\mathbf{g}$, as shown in \ref{eq:body}.
After removing this rotation-coupled gravity term, scenario \iref{abla:body} loses essential attitude information. \tref{table:ablation} demonstrates that \iref{abla:bw/og}{Body --$\mathbf{R}^T_i {}^\mathcal{G}\mathbf{g}$} causes significant performance degradation: the ATE increases by 90.0\% on EuRoC, 107.0\% on Pegasus dataset, and 360.5\% on Blackbird's Unseen sequences. 

% Interestingly, for the widely used \textbf{\iref{abla:global}{Global}} frame, encoding gravity contributes 9.6\% of the performance improvement compared to \textbf{\iref{abla:gw/og}{Global --$\mathbf{g}^\mathcal{G}$}} in Pegasus dataset. This suggests that gravity in the global frame may implicitly provide information about the ground, enhancing the network’s ability to interpret spatial dynamics.

% \paragraph{EuRoC datasets}
% In real-world quadrotor datasets EuRoC datasets, the body coordinates frame and attitude encoding consistently demonstrate superior performances, as shown in \ref{fig:euroc_ablation}. As detailed in ~\tref{euroc_ablation}, significant improvements are observed in ATE when transforming global coordinate to body coordinate and attitude encoding. Compared to the global coordinate frame, converting to body coordinate has an improvement of 53.2\% in ATE. Subsequently, encoding the attitude information leads to an additional increase of 30.3\% in ATE. Furthermore, to confirm our previously mentioned role of gravity in enhancing rotation information, we conducted experiments with body frame excluding gravity term. By removing the term, ATE increases by 90.0\% compared to the standard body frame, highlighting the important role of gravity in accuracy. 

% \subsection{Ablation study on net}

% \subsection{Rotation Noise ablation }

\section{Conclusion \& Discussion}

% We demonstrate that the body representation is more robust to the rotation error.
% In this work, we address the challenges identified in the discussion of IMO\cite{CioffiRal2023} without extra information like thrust signal. 
% We demonstrate that simply preserving body-frame IMU data, inclusive of gravitational acceleration, significantly improves IO performance on UAVs. Building upon this analysis, we propose the AirIO system, which further enhances accuracy by explicitly encoding attitude information and integrating it with an EKF. 
% Notably, compared to IMO, our model generalizes effectively to trajectories not present in the training datasets and operates without relying on additional inputs such as thrust. 
% We believe that the proposed work has broader implications for reliable state estimation in agile drone flight.

In this work, we investigate an effective representation of learning-based IO for multirotor UAVs and propose a solution that relies solely on IMU data. We identify that the commonly used global-frame representation is less effective for dynamic, agile maneuvers due to the less observable attitude information compared to IMU data represented in body-frame. 
Based on this, we develop the AirIO system by explicitly encoding attitude information and integrating it with an EKF. 
We show that our approach outperforms the state of the art \cite{CioffiRal2023} in various scenarios and environments without relying on additional sensors or control information. Our method achieves on average 66.7\%  improvement in accuracy. Furthermore, explicitly encoding attitude information into the motion network results in an additional 23.8\% improvement.
{We validate the real-time capability of AirIO on an NVIDIA Jetson AGX Orin. Future work includes deploying the system on real UAVs for closed-loop evaluation.}
% \update{Additionally, to assess the feasibility of real drone deployment, we evaluate AirIO's real-time performance on NVIDIA Jetson AGX Orin using an IMU operating at 200 Hz. AirIO achieves an average runtime of 74.23ms, demonstrating its potential for onboard use in aerial systems.}
%We believe that the proposed work has broader implications for reliable state estimation in agile drone flight.

% \clearpage
\ifCLASSOPTIONcaptionsoff
  \newpage
\fi

\bibliographystyle{IEEEtran}
% \bibliography{bibliography/papers, bibliography/yuheng}

\newpage
\appendix
\section{Appendix}

\subsection{Real-Time Performance Analysis}
\label{sec:Time Analysis}
To assess the real-time performance of AirIO, we implemented a first-in-first-out (FIFO) IMU buffer to store IMU measurements published at 200 Hz. In our setup, the buffer size is set to 1000, corresponding to approximately 5 seconds of data. At each iteration, we input the entire buffer into the AirIO and AirIMU networks, but only select the outputs corresponding to the newly received (unseen) IMU frames. After network inference, the learned velocity, IMU corrections, and associated covariances are passed to the EKF for fusion. Thus, the total inference time includes both the network computation and the EKF update.

\tref{tab:inference_time_structured} shows the comparison of real-time inference time of our approach on different platforms.  Overall, AirIO provides the state updates for 28.92 ms on low-end GPU and 74.23 ms on the Orin AGX, which is feasible for real drone deployment. We also notice that most of the inference time is spent on processing the neural network of AirIO, which highly depends on the IMU buffer size.
%By leveraging the IMU buffer, we are able to process all the IMU inputs in real time. However, during the practice, the state update will have a latency which close to the inference time of system. 
% 34.5 fps
% 13.4

% \begin{table}[h]
%     \caption{Average 1-run inference time (ms) on different platforms for EuRoC.}
%     \label{tab:inference_time_structured}
%     \centering
%     \resizebox{\linewidth}{!}{
%     \begin{tabular}{l|l|l|l|c}
%         \toprule
%         \textbf{Platform} & \textbf{CPU} & \textbf{GPU} & \textbf{OS + Environment} & \textbf{Avg. Inference Time (ms)} \\
%         \midrule
%         Host Machine & Intel Core i9-12900K & NVIDIA GeForce RTX 2060 (6GB) & Ubuntu 22.04, CUDA 12.6 & 28.92 \\
%         Jetson AGX Orin & 12-core ARM Cortex-A78AE & Ampere GPU & JetPack 3.6 & 74.23 \\
%         \bottomrule
%     \end{tabular}
%     }
% \end{table}
\begin{table}[h]
    % \normalsize
    \caption{Average 1-run inference time (ms) on different platforms for EuRoC.}
    \label{tab:inference_time_structured}
    \centering
    \resizebox{\linewidth}{!}{
    \begin{tabular}{@{}l|ll@{}}
        \toprule
        \textbf{Platform} & \textbf{Host Machine} & \textbf{Jetson AGX Orin} \\
        \midrule
        CPU & Intel Core i9-12900K & 12-core ARM Cortex-A78AE \\
        \midrule
        GPU & NVIDIA RTX 2060 (6GB) & Ampere GPU \\
        \midrule
        OS + Env & Ubuntu 22.04, CUDA 12.6 & JetPack 3.6 \\
        \midrule
        Avg. Inference Time (ms) & 28.92 & 74.23 \\
        \bottomrule
    \end{tabular}}
\end{table}

\subsection{BlackBird Dataset}
The Blackbird unmanned aerial vehicle (UAV) dataset is an indoor dataset\cite{antonini2018blackbird} designed to capture aggressive flight maneuvers for fully autonomous drone racing. It collects data using a Blackbird quadrotor platform with an Xsens MTi-3 IMU. The blackbird takes place in a motion capture room and follows a predefined periodic trajectory, each lasting approximately 3-4 minutes at high speed.
\subsubsection{Seen and Unseen sequences separation}
\label{Appendix:Seen-Unseen}
In our experiments, we define two distinct groups of trajectories: \textbf{SEEN} and \textbf{UNSEEN} sequences. SEEN sequences appear in both the training and testing phases, while UNSEEN sequences do not appear in any phase of the training process. Specifically, for SEEN sequences, we use the same five sequences that were used in IMO: \texttt{clover}, \texttt{egg}, \texttt{halfMoon}, \texttt{star}, and \texttt{winter}, with peak velocities of 5, 8, 4, 5, 4$\si{\meter\per\second}$, respectively. Each trajectory is split into training, validation, and testing sets. Since these trajectories appear in both the training and testing sets, we term them as \textbf{SEEN sequences}. To further evaluate the model's ability to adapt to unseen trajectories, we also select five additional trajectories from the Blackbird dataset: \texttt{ampersand}, \texttt{sid}, \texttt{oval}, \texttt{sphinx}, and \texttt{bentDice}, with peak velocities of 2, 5, 4, 4, 3$\si{\meter\per\second}$, respectively. Compared to the SEEN sequences, these new trajectories never appear in training or validation; therefore, we refer to them as \textbf{UNSEEN sequences}. By comparing results on both SEEN and UNSEEN sequences, we could gain a comprehensive understanding of the model's robustness and generalization capabilities.

\subsubsection{Training and Testing Sequences Separation}
In our experiments, we follow the same dataset-splitting strategy presented in IMO: for each trajectory, the data is allocated as 70\% for training, 15\% for validation, and the remaining 15\% for testing. We use the SEEN sequences' training and validation set to train our model, making our training setup identical to IMO's. For testing, we employ both the SEEN sequences' testing set and the UNSEEN sequences' testing set. The comprehensive testing setup allows us to evaluate our method's generalization to new trajectories.
\begin{table}[H]
    \normalsize
    \caption{Separation of trajectory sequences into SEEN and UNSEEN categories, and their respective allocations to training, validation, and testing sets.}
    \label{table:blacbird-split}
    \centering
    \resizebox{1\linewidth}{!}{
    \begin{tabular}{C{2.4cm}|C{1.2cm}C{1.2cm}C{1.2cm}C{1.2cm}C{1.2cm}C{1.2cm}}
\toprule
        \multirow{1}{*}
        {\textbf{SEEN}}& \multirow{1}{*}{clover}&\multirow{1}{*}{Egg}&\multirow{1}{*}{halfMoon}&\multirow{1}{*}{Star}&\multirow{1}{*}{Winter}\\

         \midrule
         training (70\%)&\ding{52}&\ding{52}&\ding{52}&\ding{52}&\ding{52}\\
         validation (15\%)&\ding{52}&\ding{52}&\ding{52}&\ding{52}&\ding{52}\\
         testing (15\%)&\ding{52}&\ding{52}&\ding{52}&\ding{52}&\ding{52}\\

\midrule
\midrule
        \multirow{1}{*}
        {\textbf{UNSEEN}}& \multirow{1}{*}{Ampersand}&\multirow{1}{*}{Sid}&\multirow{1}{*}{Oval}&\multirow{1}{*}{Sphinx}&\multirow{1}{*}{BentDice}\\ 
        training (70\%)&\ding{56}&\ding{56}&\ding{56}&\ding{56}&\ding{56}\\
         validation (15\%)&\ding{56}&\ding{56}&\ding{56}&\ding{56}&\ding{56}\\
         testing (15\%)&\ding{52}&\ding{52}&\ding{52}&\ding{52}&\ding{52}\\

        \bottomrule
    \end{tabular} 
    }
    \vspace{-10pt}
\end{table}

\subsection{Qualitative Evaluation for Blackbird dataset}
\label{Appendix:BlackBird}
\begin{figure}[h]
    \centering
    \includegraphics[width=\columnwidth]{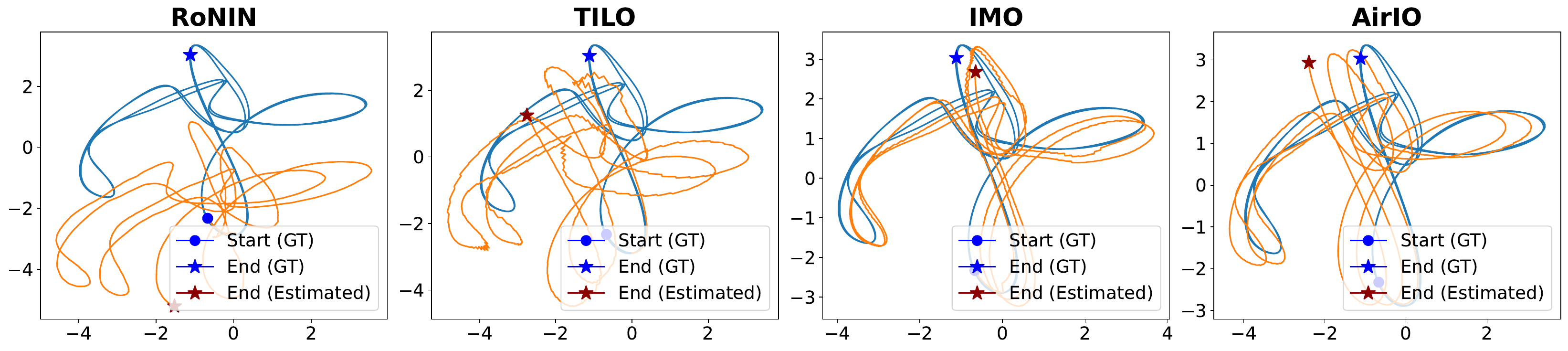}
    \vspace{0.1em}
    \includegraphics[width=\columnwidth]{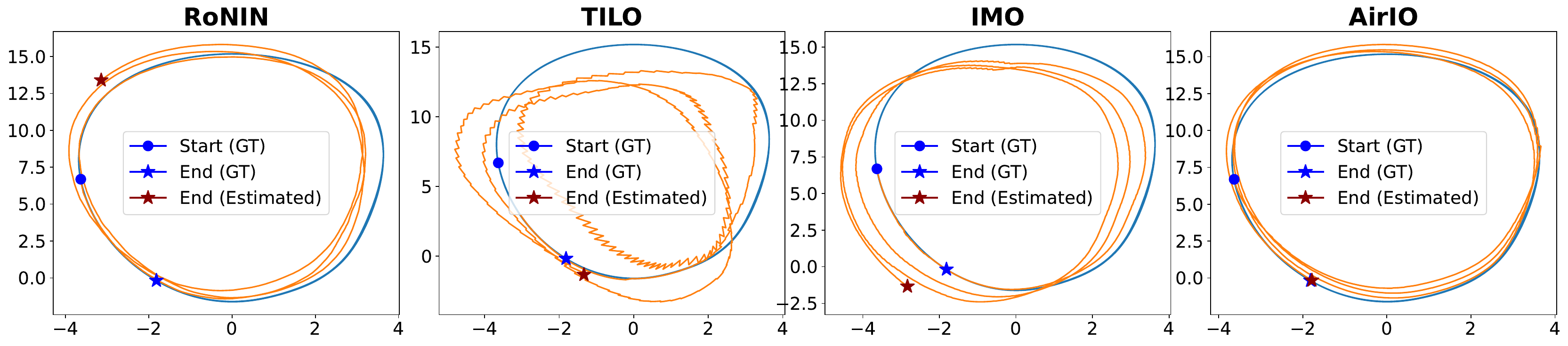}
    \vspace{0.1em}
    \includegraphics[width=\columnwidth]{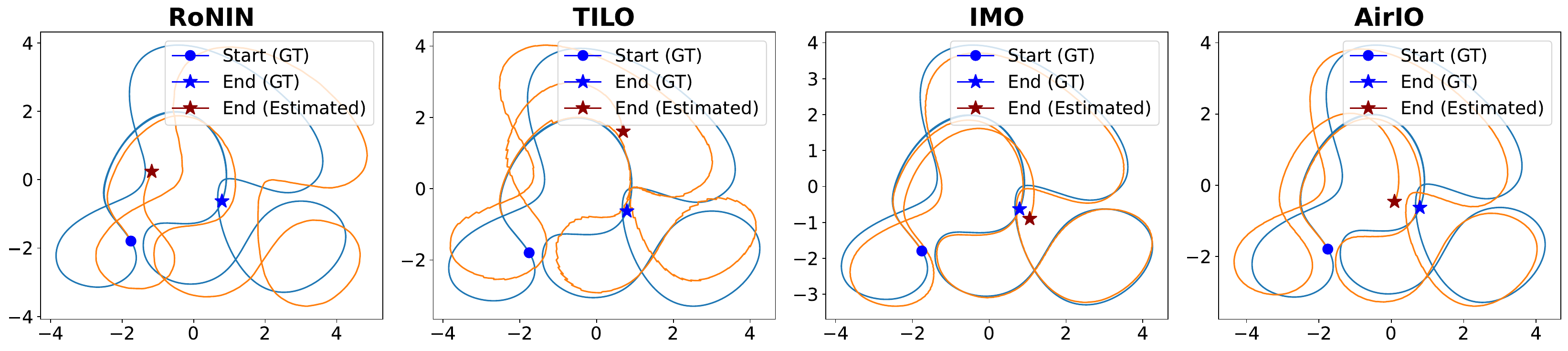}
    \vspace{0.1em}
    \includegraphics[width=\columnwidth]{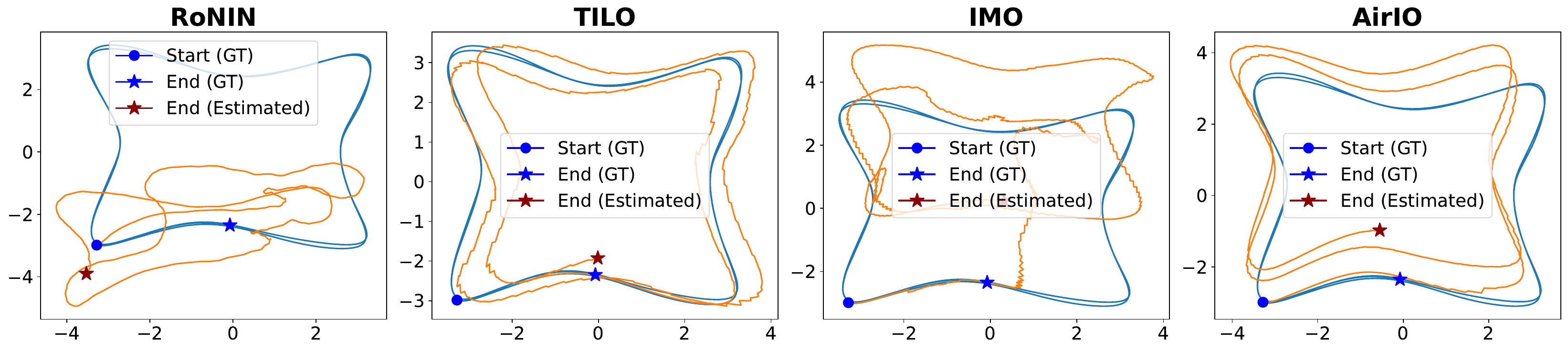}
    \caption{SEEN Sequences. From top to bottom: \texttt{Clover}, \texttt{Egg}, \texttt{Winter} and \texttt{Star}. Our method demonstrates robust performance without requiring any additional sensor information.}
    \label{fig:bb_seen}
\end{figure}

\begin{figure}[h]
    \centering
    \includegraphics[width=\columnwidth]{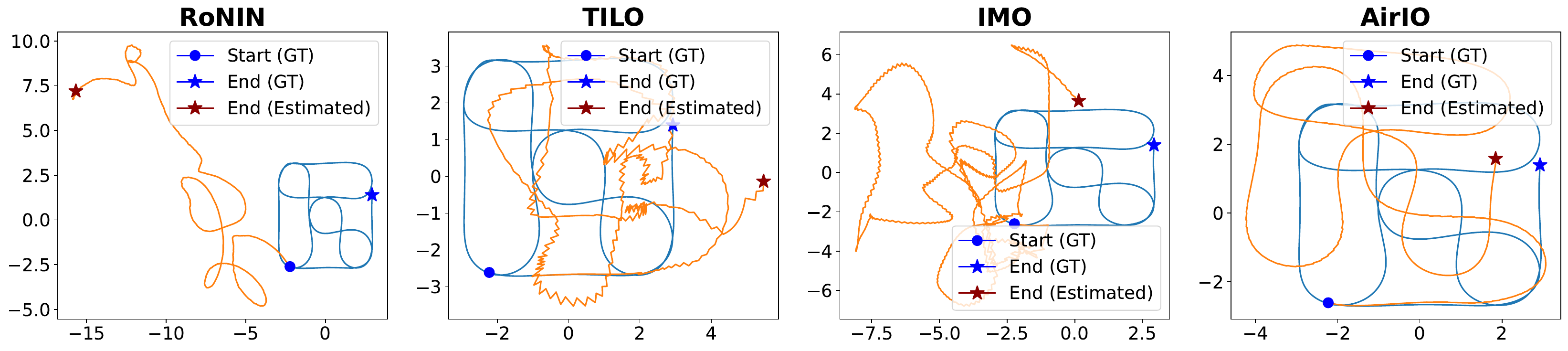}
    \vspace{0.1em}
    \includegraphics[width=\columnwidth]{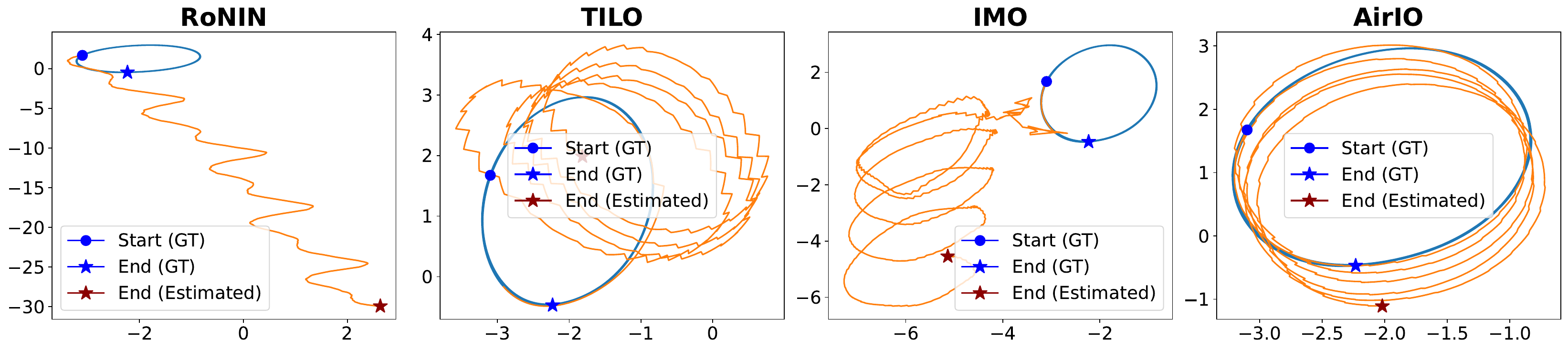}
    \vspace{0.1em}
    \includegraphics[width=\columnwidth]{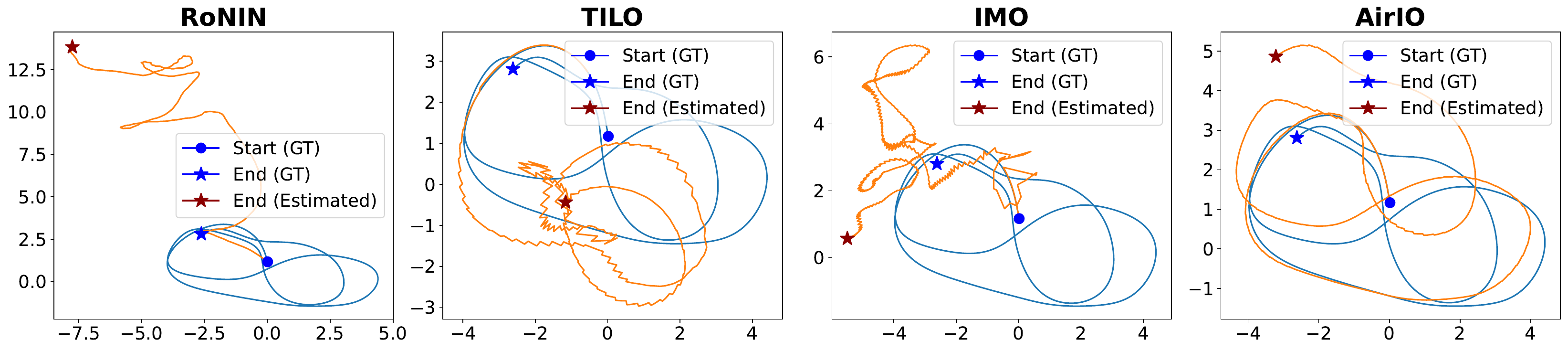}
    \caption{UNSEEN Sequences. From top to bottom: \texttt{bentDice}, \texttt{Oval}, and \texttt{Sphinx}. Our method demonstrates remarkable adaptability to trajectories it has never seen before.}
    \label{fig:bb_unseen}
\end{figure}

We present more details on the evaluation of the Blackbird dataset. As shown in \ref{fig:bb_seen} and \ref{fig:bb_unseen}, we showcase seven additional trajectories that further highlight our method's performance. Our method achieves superior performance in the SEEN sequences. For more than half of these sequences, it outperforms existing methods that rely on additional information in addition to IMU measurements, while our method uses only IMU data. When evaluating UNSEEN sequences, our model outperforms all existing methods on all sequences, demonstrating its remarkable adaptability.

% \subsection{Drone Dynamics}
% \label{Sec:QuadrotorModel}
% \begin{equation}
% \label{eq:quadrotordynamics}
% \mathbf{m} \dot v^B = {\mathbf{F}^B_{{net}_i}} = T^B_i - F^B_{{drag}_i}\mathbf - {R_i}^T \mathbf{g}^G,
% \end{equation}

% \begin{equation}
%  T^B_i = \mathbf{m} \dot v^B +  F^B_{{drag}_i}\mathbf + {R_i}^T \mathbf{g}^G,
% \end{equation}

\subsection{Pegasus Dataset}
\label{Appendix:Pegasus_Dataset}
We collected a simulation dataset in the open-source Pegasus Simulator \cite{Jacinto2024pegasus} to evaluate our proposed method under controlled conditions. Pegasus is a framework built on top of NVIDIA Omniverse and Isaac Sim. It is designed for multirotor simulation and supports integration with PX4 firmware, as well as Python control interfaces. In our setup, we used QGroundControl to control the multirotor and also employed the quadratic thrust curve and linear drag model, ensuring the generated flight dynamics closely resemble real-world conditions.

\begin{figure}[h]
    \centering
    \begin{subfigure}[b]{0.48\columnwidth}
        \centering
        \includegraphics[width=\textwidth]{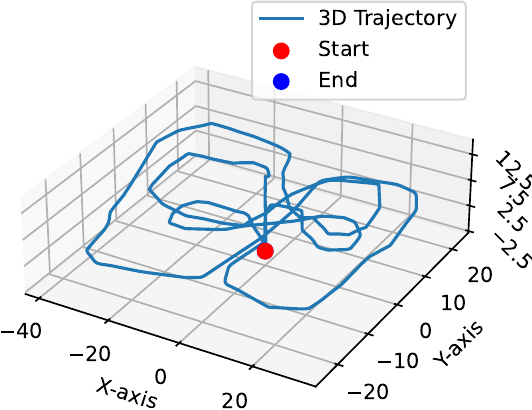}
        \caption{TEST\_1: This trajectory covers a  distance of 558.6$\meter$ and a total duration of 253.2$\second$, with a peak speed of 4.7 $\si{\meter\per\second}$, an average speed of 2.2 $\si{\meter\per\second}$}
        \label{fig:sub_test1}
    \end{subfigure}
    \hfill
    \begin{subfigure}[b]{0.48\columnwidth}
        \centering
        \includegraphics[width=\textwidth]{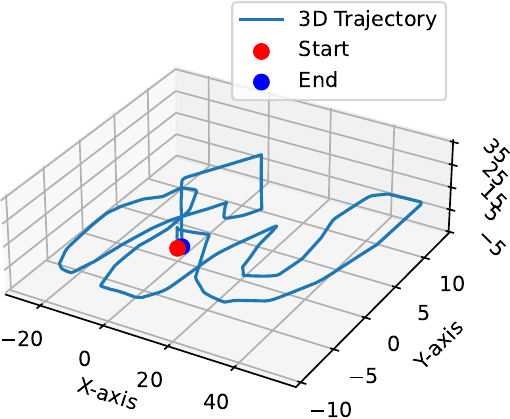}
        \caption{TEST\_2: This trajectory covers a  distance of 316.7$\meter$ and a total duration of 228.5$\second$, with a peak speed of 4.6 $\si{\meter\per\second}$, an average speed of 1.4 $\si{\meter\per\second}$}
        \label{fig:sub_test2}
    \end{subfigure}

     \vspace{1em}
         \begin{subfigure}[b]{0.48\columnwidth}
        \centering
        \includegraphics[width=\textwidth]{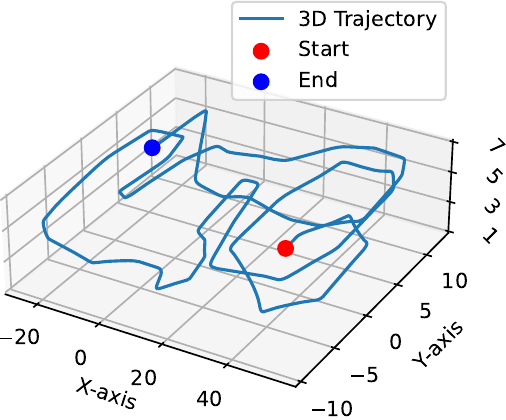}
        \caption{TEST\_3: This trajectory covers a  distance of 402.1$\meter$ and a total duration of 148.8$\second$, with a peak speed of 4.9 $\si{\meter\per\second}$, an average speed of 2.7 $\si{\meter\per\second}$}
        \label{fig:sub_test3}
    \end{subfigure}
    \caption{Testing set}
    \label{fig:testmain}

\end{figure}

In our experiment, we collected a total of seven trajectories datasets, named \textbf{Pegasus Dataset}. We divided the Pegasus dataset into training and testing sets. Specifically, four trajectories are selected for training and the remaining three trajectories are testing. They are illustrated in \ref{fig:trainmain} and \ref{fig:testmain}. We provide a detailed overview of each trajectory as follows.
\begin{figure}[h]
    \centering
    \begin{subfigure}[b]{0.48\columnwidth}
        \centering
        \includegraphics[width=\textwidth]{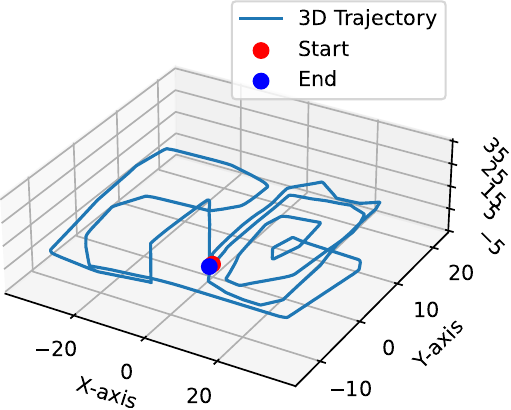}
        \caption{TRAIN\_1: Distance 516.2$\meter$, Duration 254.7$\second$, Peak 4.8, Avg. 2.0 $\si{\meter\per\second}$.}
        \label{fig:sub_train1}
    \end{subfigure}
    \hfill
    \begin{subfigure}[b]{0.48\columnwidth}
        \centering
        \includegraphics[width=\textwidth]{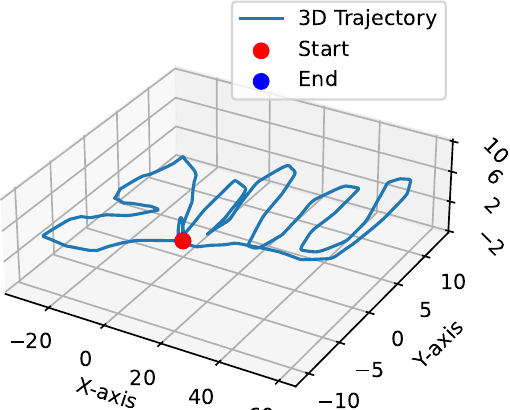}
        \caption{TRAIN\_2: Distance 329.0$\meter$, Duration 165.8$\second$, Peak 4.3, Avg. 2.0 $\si{\meter\per\second}$.}
        \label{fig:sub_train2}
    \end{subfigure}

    \vspace{1em}
    \begin{subfigure}[b]{0.48\columnwidth}
        \centering
        \includegraphics[width=\textwidth]{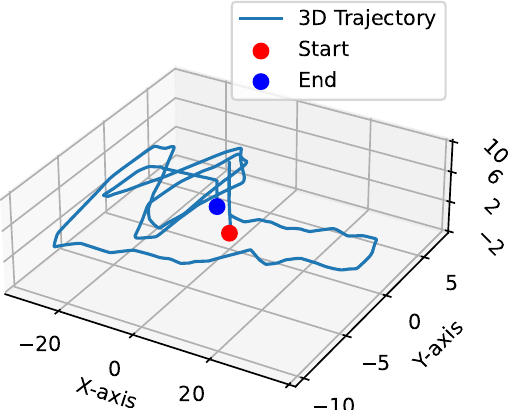}
        \caption{TRAIN\_3: Distance 263.5$\meter$, Duration 355.5$\second$, Peak 4.6, Avg. 0.7 $\si{\meter\per\second}$.}
        \label{fig:sub_train3}
    \end{subfigure}
    \hfill
    \begin{subfigure}[b]{0.48\columnwidth}
        \centering
        \includegraphics[width=\textwidth]{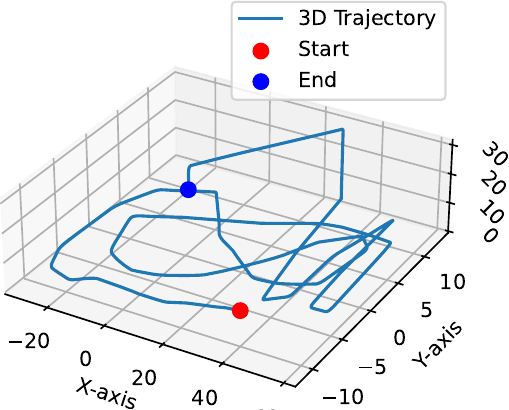}
        \caption{TRAIN\_4: Distance 452.1$\meter$, Duration 160.4$\second$, Peak 4.9, Avg. 2.8 $\si{\meter\per\second}$.}
        \label{fig:sub_train4}
    \end{subfigure}
    \caption{Training trajectories (1--4).}
    \label{trainmain}
\end{figure}

\subsection{EuRoC Dataset}
The EuRoC datasets are the well-known benchmarks for odometry and SLAM algorithms. They are collected by a micro aerial vehicle: an AscTec Firefly hex-rotor helicopter. There are 11 trajectories collected in two scenarios: an industrial environment and a motion capture room. We selected \texttt{MH\_01\_easy}, \texttt{MH\_03\_medium}, \texttt{MH\_05\_difficult}, \texttt{V1\_02\_medium}, \texttt{V2\_01\_easy}, \texttt{V2\_03\_difficult} for training, and the rest for testing.
\begin{figure}[H]
    \centering
    \includegraphics[width=1\linewidth]{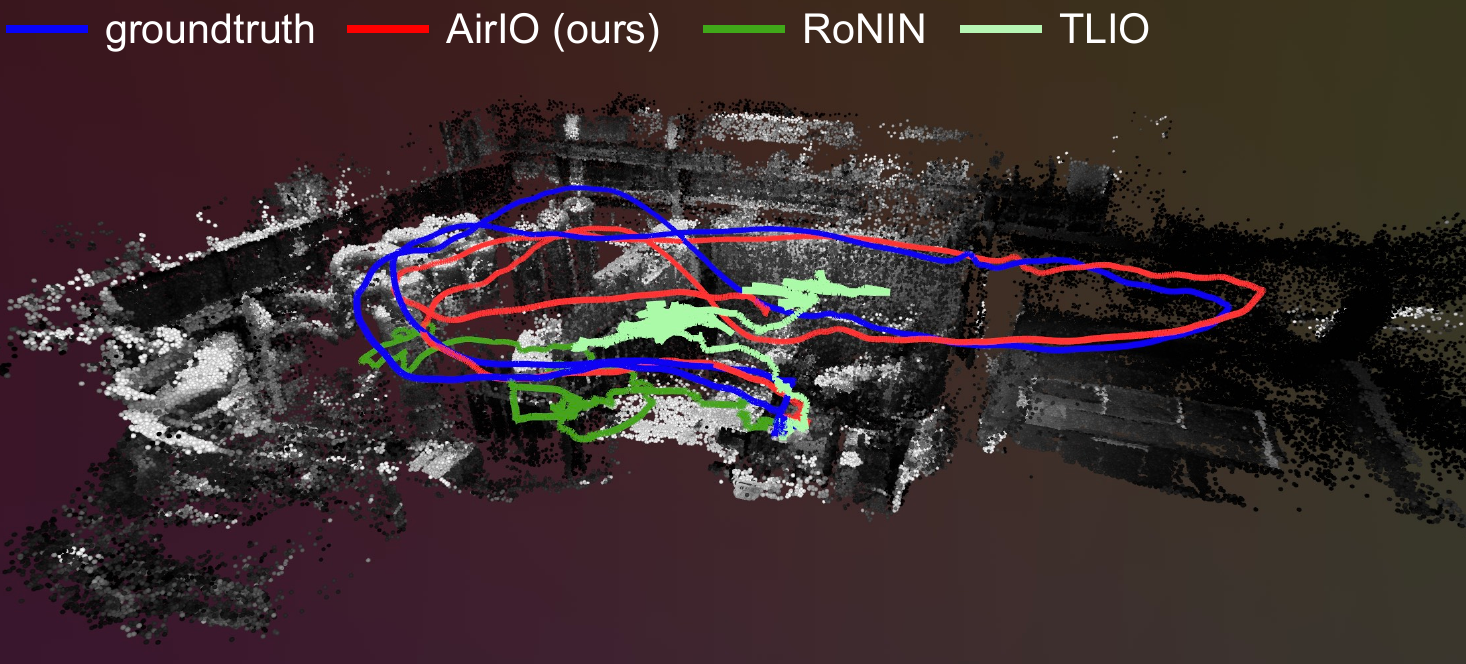}
    \caption{The MH\_04\_difficult trajectories from the EuRoC dataset visualized within its 3D reconstruction map. While RoNIN (dark green) and TLIO (light green) fail, AirIO (red) retains a coherent trajectory shape.  }
    \label{fig:mh04_reconstruction}
\end{figure}

\subsection{Ablation Study in Pegasus and EuRoC dataset}
\label{Appendix:Pegasus_eval}
\begin{table}[H]
    \normalsize
    \caption{Ablation study on the EuRoC and Pegasus datasets comparing different feature representations. Evaluation metric: RTE (Unit: $\meter$).}
    \label{table:ablation_rte}
    \centering
    \resizebox{1\linewidth}{!}{
    \begin{tabular}{C{1.8cm}|C{1.25cm}C{1.25cm}C{1.25cm}C{1.25cm}C{1.25cm}C{1.25cm}C{1.25cm}}
\toprule
        \multirow{2}{*}
        {\textbf{Seq.}}& \multirow{2}{*}{\shortstack{\textbf{Global} \\ \textbf{$-$ gravity}}}& \multirow{2}{*}{\textbf{Global}}& \multirow{2}{*}{\shortstack{\textbf{Body} \\ \textbf{$-$ gravity}}}& \multirow{2}{*}{\shortstack{\textbf{Global} \\ \textbf{$+$Attitude}}} & \multirow{2}{*}{\textbf{Body}} & \multirow{2}{*}{\shortstack{\textbf{Body} \\ \textbf{$+$Attitude}}}\\ 
        & & & & & & \\

         \midrule
        \raggedright \textbf{EuRoC}  & \multicolumn{1}{c}{} & \multicolumn{1}{c}{} & \multicolumn{1}{c}{} & \multicolumn{1}{c}{} & \multicolumn{1}{c}{} & \multicolumn{1}{c}{} \\
   
         MH02& \cellcolor{g6!100} 1.684 & \cellcolor{g4!100} 1.575 & \cellcolor{g7!100} 2.346 & \cellcolor{g3!100} 1.542 & \cellcolor{g2!100} 1.314 & \cellcolor{g1!100} \textbf{0.972} \\
         MH04 &\cellcolor{g7!100} 2.618 & \cellcolor{g4!100} 1.961 & \cellcolor{g6!100} 2.525 & \cellcolor{g3!100} 1.707 & \cellcolor{g2!100} 1.329 & \cellcolor{g1!100} \textbf{1.009} \\
         
         V103 &\cellcolor{g2!100} 1.407 & \cellcolor{g1!100} \textbf{1.352} & \cellcolor{g6!100} 1.613 & \cellcolor{g3!100} 1.485 & \cellcolor{g7!100} 1.623 & \cellcolor{g4!100} 1.512 \\
         
         V202 &\cellcolor{g3!100} 1.721 & \cellcolor{g6!100} 1.789 & \cellcolor{g7!100} 2.176 & \cellcolor{g4!100} 1.723 & \cellcolor{g2!100} 1.373 & \cellcolor{g1!100} \textbf{1.263} \\
         V101  & \cellcolor{g6!100} 1.801 & \cellcolor{g7!100} 1.991 & \cellcolor{g3!100} 1.463 & \cellcolor{g4!100} 1.498 & \cellcolor{g2!100} 1.122 & \cellcolor{g1!100} \textbf{1.104} \\
         
         Avg. &\cellcolor{g6!100} 1.846 & \cellcolor{g4!100} 1.734 & \cellcolor{g7!100} 2.025 & \cellcolor{g3!100} 1.591 & \cellcolor{g2!100} 1.352 & \cellcolor{g1!100} \textbf{1.172} \\

\midrule
\midrule
        \raggedright \textbf{Pegasus}  & \multicolumn{1}{c}{} & \multicolumn{1}{c}{} & \multicolumn{1}{c}{} & \multicolumn{1}{c}{} & \multicolumn{1}{c}{}  \\     
        TEST\_1& \cellcolor{g6!100} 2.783 & \cellcolor{g7!100} 2.971 & \cellcolor{g4!100} 2.134 & \cellcolor{g3!100} 1.694 & \cellcolor{g2!100} 1.137 & \cellcolor{g1!100} \textbf{1.017} \\
        TEST\_2 & \cellcolor{g6!100} 2.704 & \cellcolor{g7!100} 3.007 & \cellcolor{g2!100} 1.961 & \cellcolor{g4!100} 2.339 & \cellcolor{g3!100} 2.162 & \cellcolor{g1!100} \textbf{1.905} \\
        TEST\_3 & \cellcolor{g6!100} 3.274 & \cellcolor{g7!100} 3.350 & \cellcolor{g3!100} 1.298 & \cellcolor{g4!100} 1.969 & \cellcolor{g2!100} 0.584 & \cellcolor{g1!100} \textbf{0.396} \\
        Avg. & \cellcolor{g6!100} 2.920 & \cellcolor{g7!100} 3.109 & \cellcolor{g3!100} 1.798 & \cellcolor{g4!100} 2.001 & \cellcolor{g2!100} 1.294 & \cellcolor{g1!100} \textbf{1.106} \\
        
\midrule
\midrule
        \raggedright \textbf{Blackbird}  & \multicolumn{1}{c}{} & \multicolumn{1}{c}{} & \multicolumn{1}{c}{} & \multicolumn{1}{c}{} & \multicolumn{1}{c}{}  \\ 
          Ampersand &\cellcolor{g6!100} 8.933 & \cellcolor{g3!100} 6.845 & \cellcolor{g7!100} 11.379 & \cellcolor{g4!100} 8.374 & \cellcolor{g2!100} 1.254 & \cellcolor{g1!100} \textbf{1.185} \\
        Sid &\cellcolor{g7!100} 7.768 & \cellcolor{g6!100} 6.151 & \cellcolor{g4!100} 3.575 & \cellcolor{g3!100} 2.72 & \cellcolor{g2!100} 0.737 & \cellcolor{g1!100} \textbf{0.504} \\
        Oval&\cellcolor{g6!100} 4.154 & \cellcolor{g4!100} 3.936 & \cellcolor{g3!100} 1.094 & \cellcolor{g7!100} 4.555 & \cellcolor{g2!100} 0.690 & \cellcolor{g1!100} \textbf{0.558} \\
        Sphinx&\cellcolor{g7!100} 4.637 & \cellcolor{g6!100} 3.647 & \cellcolor{g3!100} 1.695 & \cellcolor{g4!100} 2.875 & \cellcolor{g1!100} \textbf{0.888} & \cellcolor{g2!100} 1.021 \\
        BentDice&\cellcolor{g7!100} 6.998 & \cellcolor{g6!100} 5.587 & \cellcolor{g3!100} 5.427 & \cellcolor{g4!100} 5.578 & \cellcolor{g2!100} 1.389 & \cellcolor{g1!100} \textbf{0.871} \\
        Avg.& \cellcolor{g7!100} 6.498 & \cellcolor{g6!100} 5.233 & \cellcolor{g3!100} 4.634 & \cellcolor{g4!100} 4.82 & \cellcolor{g2!100} 0.992 & \cellcolor{g1!100} \textbf{0.828} \\
        \bottomrule
        \bottomrule
    \end{tabular} 
    }
    \vspace{-10pt}
\end{table}

\subsection{Ablation Study on Model Compression}
\label{Appendix:Model Size}
To evaluate the compressibility of different representations, we introduced additional two lightweight models \textbf{Light A} and \textbf{Light B}. The light models keep the same layer structure but shrink the dimensions of each layer's hidden units. Finally, the encoder's latent feature dimension is reduced from 256 to 128, and then further to 64, yielding progressively smaller models.

To quantify the performance degradation as the model is compressed, we define the degradation ratio for ATE and RTE. A higher degradation ratio indicates a larger drop in performance. As shown in \ref{table:ablation_model}, the model under body frame shows smoother degradation in both ATE and RTE.
\begin{table}[H]
    \normalsize
    \caption{Ablation study on the Blackbird, Pegasus, and EuROC datasets comparing compressibility of models under body frame and global frame. Evaluation metric: ATE (Unit: $\meter$), RTE(Unit: $\meter$), and Degradation.}
    \label{table:ablation_model}
    \centering
    \resizebox{0.6\linewidth}{!}{
\begin{tabular}{C{1.4cm}C{1.4cm}|C{1cm}C{1.7cm}|C{1cm}C{1.7cm}|C{1cm}C{1.7cm}}
\toprule
\multicolumn{2}{c|}{\textbf{Model}} & \multicolumn{2}{c|}{\textbf{Regular}}&\multicolumn{2}{c|}{\textbf{Light A}}&\multicolumn{2}{c}{\textbf{Light B}}  \\
 \midrule
 \multicolumn{2}{c|}{\textbf{Feature Size}} & \multicolumn{2}{c|}{256$\times$1}&\multicolumn{2}{c|}{128$\times$1}&\multicolumn{2}{c}{64$\times$1}  \\
\midrule
 \multicolumn{2}{c|}{\textbf{Model Size}} & \multicolumn{2}{c|}{2.524 MB}&\multicolumn{2}{c|}{0.641 MB}&\multicolumn{2}{c}{0.175 MB}  \\
\bottomrule
 \toprule
\multicolumn{2}{c|}{\textbf{Metrics}} & ATE &Degradation & ATE &Degradation&ATE &Degradation \\
 \midrule
\multirow{2}{*}{\textbf{Blackbird}}&Body&0.647&-&0.755&\cellcolor{g2!100}16.8\%&0.931&\cellcolor{g2!100}44.0\%\\
&Global&0.837&-&1.238&47.9\%&1.522&	81.8\%\\
 \midrule
  \midrule
 \multirow{2}{*}{\textbf{Pegasus}}&Body& 4.670&-&10.118&116.6\%&15.192&\cellcolor{g2!100}225.3\%\\
&Global& 17.278&-&30.950&\cellcolor{g2!100}79.1\%&69.236&300.7\% \\
 \midrule
  \midrule
  \multirow{2}{*}{\textbf{EuRoC}}&Body&4.730&-&5.447&\cellcolor{g2!100}15.2\%&6.875&	\cellcolor{g2!100}45.4\%\\
&Global&10.096&-&14.033&39.0\%&38.236&278.7\%\\
\bottomrule

 \toprule
\multicolumn{2}{c|}{\textbf{Metrics}} & RTE &Degradation & RTE &Degradation&RTE &Degradation \\
 \midrule
\multirow{2}{*}{\textbf{Blackbird}}&Body&0.345&-&0.454& 31.3\%&0.510&\cellcolor{g2!100}47.7\%	\\
&Global&0.525&-&0.583&\cellcolor{g2!100}11.0\%&0.983&87.1\%\\
 \midrule
  \midrule
\multirow{2}{*}{\textbf{Pegasus}}&Body&1.516&-&2.226&46.9\%&2.203&\cellcolor{g2!100}45.3\%\\
&Global&3.109&-&3.422&\cellcolor{g2!100}10.0\%&4.858&56.2\%\\
 \midrule
  \midrule
\multirow{2}{*}{\textbf{EuRoC}}&Body&1.352&-&1.359&\cellcolor{g2!100}0.5\%&1.297&\cellcolor{g2!100}-4.1\%\\
&Global&1.734&-&2.176&25.5\%&4.468&157.8\%\\
 \midrule

\end{tabular}
    }

\end{table}

\subsection{Comparison with Visual Inertial Odometry}
\label{Appendix:VIO}
We have added the results of OpenVINS~\cite{geneva2020openvins} as a reference. Specifically, we incorporate the quantitative results reported by IMO~\cite{CioffiRal2023} for the BlackBird dataset and by QVIO~\cite{peng2024quantized} for the EuRoC dataset.

As shown in the Table. \ref{openvins-appendix}, on the EuRoC dataset, the VIO system exhibits a significant advantage over IO methods due to the additional vision modality, achieving a 64.1\% improvement in ATE. However, on the BlackBird dataset, our method achieves comparable performance across all sequences. Notably, Blackbird dataset features peak velocities up to 8~$\mathrm{m/s}$, leading to motion blur and large optical flows if using vision-based approaches, which make feature tracking challenging and degrade the performance of VIO systems.

\begin{table}[h]
    \caption{The ATE (Unit: \meter) and RTE (Unit: \meter) on the Blackbird dataset and the EuRoC dataset. \textbf{Seen} are sequences where the training and testing datasets are derived from the same trajectory.}
    \label{openvins-appendix}
    \centering
    \resizebox{1\linewidth}{!}{
    \begin{tabular}{C{2.5cm}|C{.8cm}C{.8cm}|C{.8cm}C{.8cm}C{.8cm}C{.8cm}C{.8cm}C{.8cm}C{.8cm}C{.8cm}}
        \toprule
       \multirow{2}{*}{\textbf{Seq.}} & \multicolumn{2}{c}{\textbf{OpenVINS$^\S$}~\cite{geneva2020openvins}}&   \multicolumn{2}{c}{\textbf{TLIO$^\dagger$}} & \multicolumn{2}{c}{\textbf{IMO$^\dagger$$^\ast$ }} & \multicolumn{2}{c}{\textbf{AirIO Net}} & \multicolumn{2}{c}{\textbf{AirIO EKF}}\\
         &  RTE&  ATE  &  RTE &  ATE  &  RTE &  ATE  &  RTE &  ATE  &  RTE &  ATE \\
        \midrule
        \raggedright \textbf{EuRoC} & \multicolumn{2}{c}{} & \multicolumn{2}{c}{} & \multicolumn{2}{c}{} & \multicolumn{2}{c}{} &\\
         MH02 &  - &\cellcolor{g1!100} \textbf{0.9} & \cellcolor{g5!100} 2.121 & \cellcolor{g5!100} 5.902 & \cellcolor{g7!100} 2.451 & \cellcolor{g7!100} 7.281 & \cellcolor{g1!100} \textbf{0.936} & \cellcolor{g3!100} 4.917 & \cellcolor{g3!100} 0.987 & \cellcolor{g2!100} 2.478 \\
         
         MH04 &-&\cellcolor{g1!100} \textbf{1.36} & \cellcolor{g5!100} 4.542 & \cellcolor{g5!100} 8.586 & \cellcolor{g7!100} 5.498 & \cellcolor{g7!100} 8.626 & \cellcolor{g3!100} 1.093 & \cellcolor{g3!100} 2.726 & \cellcolor{g1!100} \textbf{1.005} & \cellcolor{g2!100} 2.308\\

         V103 &
 - &\cellcolor{g1!100} \textbf{1.68} & \cellcolor{g5!100} 1.695 & \cellcolor{g3!100} 3.24 & \cellcolor{g7!100} 2.58 & \cellcolor{g7!100} 7.863 & \cellcolor{g1!100} \textbf{1.519} & \cellcolor{g5!100} 3.844 & \cellcolor{g3!100} 1.552 & \cellcolor{g2!100} 3.05 \\

         V202 & - &\cellcolor{g1!100} \textbf{1.24} & \cellcolor{g5!100} 2.303 & \cellcolor{g7!100} 7.445 & \cellcolor{g7!100} 2.783 & \cellcolor{g5!100} 6.26 & \cellcolor{g1!100} \textbf{1.303} & \cellcolor{g3!100} 4.823 & \cellcolor{g3!100} 1.313 & \cellcolor{g2!100} 4.206 \\
         
         V101 & - &\cellcolor{g1!100} \textbf{0.53} & \cellcolor{g5!100} 1.918 & \cellcolor{g7!100} 8.576 & \cellcolor{g7!100} 1.946 & \cellcolor{g5!100} 4.814 & \cellcolor{g3!100} 1.137 & \cellcolor{g2!100} 2.917 & \cellcolor{g1!100} \textbf{1.103} & \cellcolor{g3!100} 3.844 \\
        \midrule
        \textbf{Avg.} & - &\cellcolor{g1!100} \textbf{1.14} & \cellcolor{g5!100} 2.516 & \cellcolor{g5!100} 6.75 & \cellcolor{g7!100} 3.052 & \cellcolor{g7!100} 6.969 & \cellcolor{g3!100} 1.198 & \cellcolor{g3!100} 3.846 & \cellcolor{g1!100} \textbf{1.192} &\cellcolor{g2!100} 3.177\\
 
        \midrule
        \midrule
        \raggedright \textbf{BlackBird Seen} & \multicolumn{2}{c}{} & \multicolumn{2}{c}{} & \multicolumn{2}{c}{} & \multicolumn{2}{c}{} &\\
        
        \raggedright\hspace{0.3cm} Clover &- & \cellcolor{g5!100} 0.5 & \cellcolor{g7!100} 0.797 & \cellcolor{g7!100} 1.464 & \cellcolor{g5!100} 0.681 & \cellcolor{g2!100} 0.381 & \cellcolor{g1!100} \textbf{0.368} & \cellcolor{g3!100} 0.434 & \cellcolor{g3!100} 0.391 & \cellcolor{g1!100} \textbf{0.367} \\
 
        \raggedright\hspace{0.3cm} Egg &- & \cellcolor{g3!100} 1.07 & \cellcolor{g7!100} 2.398 & \cellcolor{g7!100} 2.227 & \cellcolor{g5!100} 0.828 & \cellcolor{g5!100} 1.153 & \cellcolor{g3!100} 0.391 & \cellcolor{g2!100} 0.713 & \cellcolor{g1!100} \textbf{0.344} & \cellcolor{g1!100} \textbf{0.408} \\

        \raggedright\hspace{0.3cm} halfMoon &-& \cellcolor{g1!100} \textbf{0.37} & \cellcolor{g7!100} 0.475 & \cellcolor{g7!100} 0.956 & \cellcolor{g1!100} \textbf{0.24} & \cellcolor{g5!100} 0.761 & \cellcolor{g5!100} 0.256 & \cellcolor{g3!100} 0.491 & \cellcolor{g3!100} 0.253 & \cellcolor{g2!100} 0.457 \\
        
        \raggedright\hspace{0.3cm} Star &- & \cellcolor{g7!100} 2.78 & \cellcolor{g5!100} 0.784 & \cellcolor{g3!100} 0.68 & \cellcolor{g7!100} 3.066 & \cellcolor{g5!100} 2.13 & \cellcolor{g3!100} 0.401 & \cellcolor{g1!100} \textbf{0.442} & \cellcolor{g1!100} \textbf{0.341} & \cellcolor{g2!100} 0.477 \\
        
        \raggedright\hspace{0.3cm} Winter &-& \cellcolor{g1!100} \textbf{0.12} & \cellcolor{g7!100} 0.755 & \cellcolor{g7!100} 0.616 & \cellcolor{g5!100} 0.206 & \cellcolor{g2!100} 0.219 & \cellcolor{g1!100} \textbf{0.147} & \cellcolor{g5!100} 0.348 & \cellcolor{g3!100} 0.164 & \cellcolor{g3!100} 0.307 \\
        \midrule
        \textbf{Avg.} & 
- & \cellcolor{g5!100} 0.97 & \cellcolor{g7!100} 1.042 & \cellcolor{g7!100} 1.189 & \cellcolor{g5!100} 1.004 & \cellcolor{g3!100} 0.929 & \cellcolor{g3!100} 0.312 & \cellcolor{g2!100} 0.486 & \cellcolor{g1!100} \textbf{0.299} & \cellcolor{g1!100} \textbf{0.403} \\
    
        \bottomrule
        \multicolumn{9}{l}{$^\dagger$ Leverage ground truth orientation to transform the input data for inference.}\\
        \multicolumn{9}{l}{$^\S$ Visual-Inertial Odometry (VIO) method.}\\
    \end{tabular} 
    }
\end{table}

\end{document}